\documentclass[10pt,twocolumn,letterpaper]{article}

\usepackage{cvpr}
\usepackage{times}
\usepackage{epsfig}
\usepackage{graphicx}
\usepackage{amsmath}
\usepackage{amssymb}

\usepackage{subfigure}

\usepackage{amsmath}
\usepackage{amssymb}
\usepackage{mathrsfs}
\usepackage{bm}

\usepackage{url}
\usepackage{multirow}
\usepackage{floatrow}
\newfloatcommand{capbtabbox}{table}[][\linewidth]

\usepackage[breaklinks=true,bookmarks=false]{hyperref}

\cvprfinalcopy 

\def\cvprPaperID{5237} 

\ifcvprfinal\fi
\begin{document}

\title{Exemplar Normalization for Learning Deep Representation}

\author{Ruimao Zhang$^{1}\thanks{Equal contribution}$~,~ Zhanglin Peng$^{1*}$,~ Lingyun Wu$^1$, ~Zhen Li$^{3,4}$, ~Ping Luo$^2$ \\
$^1$ SenseTime Research, $^2$ The University of Hong Kong, \\
$^3$ The Chinese University of Hong Kong (Shenzhen), $^4$ Shenzhen Research Institute of Big Data \\
{\tt\small \{zhangruimao, pengzhanglin, wulingyun\}@sensetime.com, lizhen@cuhk.edu.cn,  pluo.lhi@gmail.com }
}


\maketitle
\thispagestyle{empty}

\begin{abstract}
 Normalization techniques are important in different advanced neural networks and different tasks.
%
This work investigates a novel dynamic learning-to-normalize (L2N) problem by proposing Exemplar Normalization (EN), which is able to learn different normalization methods for different convolutional layers and image samples of a deep network. EN significantly improves flexibility of the recently proposed switchable normalization (SN), which solves a static L2N problem by linearly combining several normalizers in each normalization layer (the combination is the same for all samples).
Instead of directly employing a multi-layer perceptron (MLP) to learn data-dependant parameters as conditional batch normalization (cBN) did, the internal architecture of EN is carefully designed to stabilize its optimization, leading to many appealing benefits.
%
%
%
%
%
(1) EN enables different convolutional layers, image samples, categories, benchmarks, and tasks to use different normalization methods, shedding light on analyzing them in a holistic view.
%
(2) EN is effective for various network architectures and tasks.
(3) It could replace any normalization layers in a deep network and still produce stable model training.
Extensive experiments demonstrate the effectiveness of EN in wide spectrum of tasks including image recognition, noisy label learning, and semantic segmentation.
For example, by replacing BN in the ordinary ResNet50, improvement produced by EN is 300\% more than that of SN on both ImageNet and the noisy WebVision dataset.
The codes and models will be released.
\end{abstract}

\section{Introduction}

\begin{figure}[htbp]
\centering
\subfigure[The learning dynamic of EN ratios of four categories in three layers.]{
\begin{minipage}[t]{1\linewidth}
\centering
\includegraphics[width=1\linewidth]{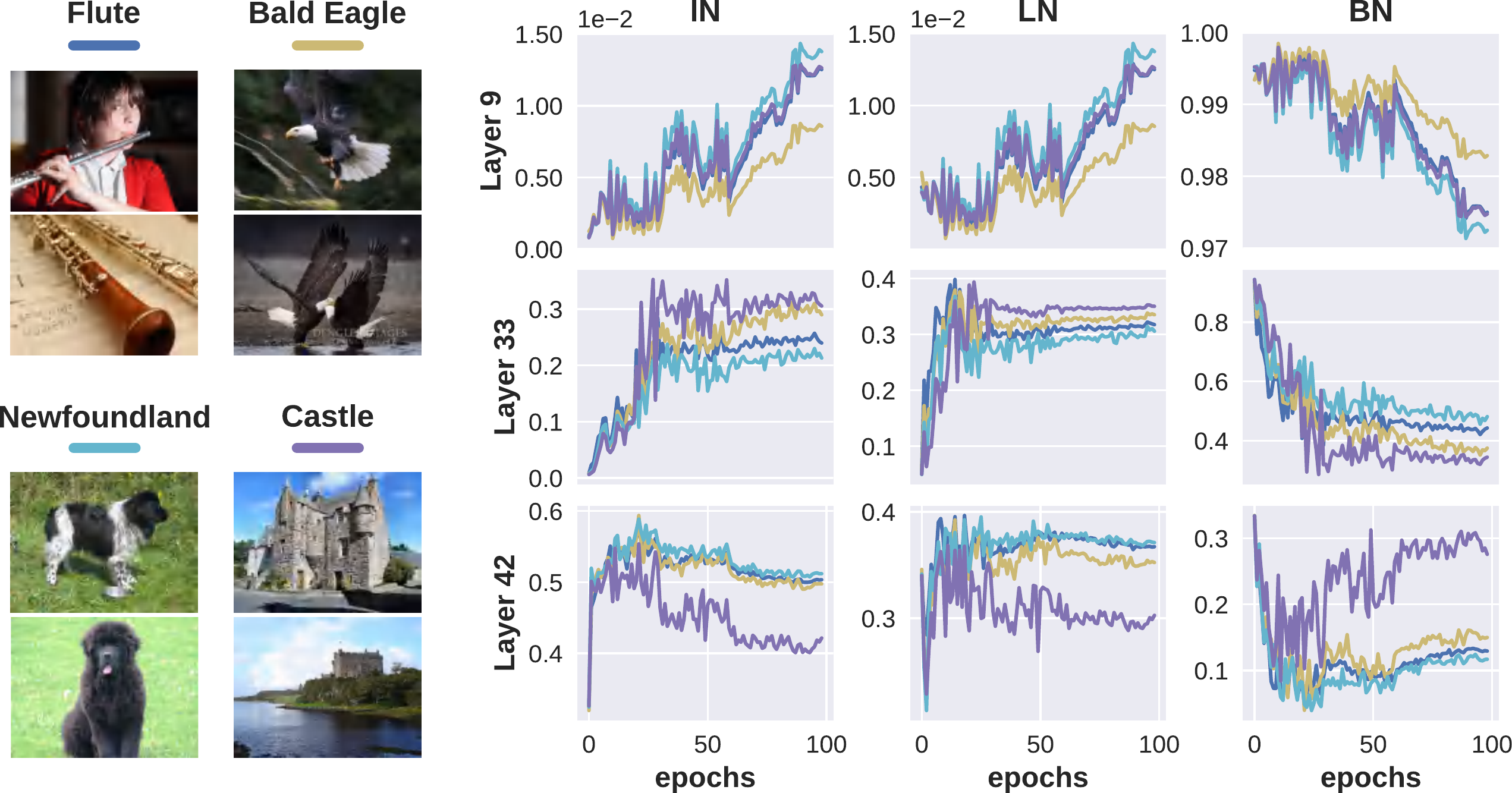}
\label{fig:fig1ratio}
\end{minipage}
} \\
\subfigure[Performance of EN and its counterparts on various CV tasks.]{
\begin{minipage}[t]{1.0\linewidth}
\centering
\includegraphics[width=1\linewidth]{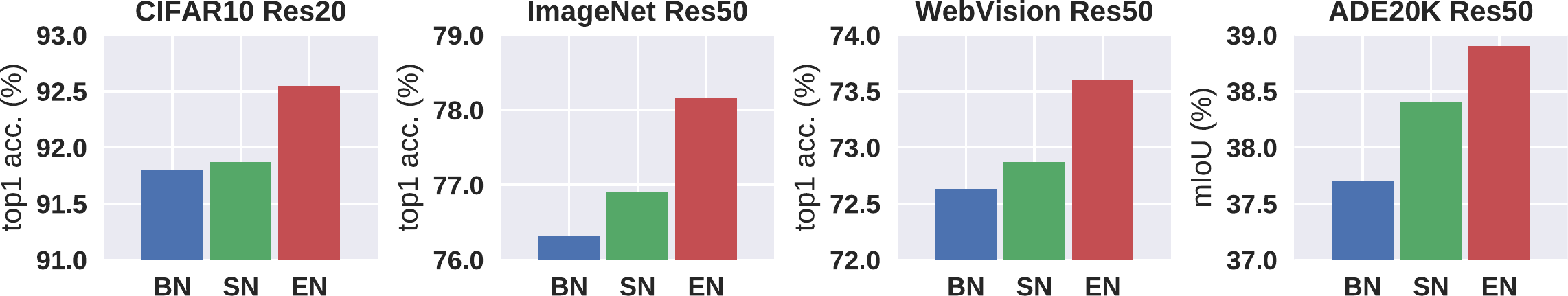}
\label{fig:fig1acc}
\end{minipage}%
}
\centering
\caption{ (a) The proposed Exemplar Normalization (EN) enables different categories to learn to select different normalizers in different layers. The four categories of ImageNet (\ie flute, bald eagle, newfoundland and castle) in three layers (\ie bottom, middle and top) of ResNet50 are presented.
(b) EN outperforms its counterparts on various computer vision tasks ( \ie image classification, noisy-supervised classification and semantic image segmentation ) by using different network architectures.
Zoom in three times for the best view.}
\label{fig:figure1}
\end{figure}

Normalization techniques are one of the most essential components to improve performance and accelerate training of convolutional neural networks (CNNs).
Recently, a family of normalization methods is proposed
including batch normalization (BN)~\cite{C:BN}, instance normalization (IN)~\cite{A:IN}, layer normalization (LN)~\cite{A:LN} and group normalization (GN)~\cite{C:GN}.
As these methods were designed for different tasks, they often normalize feature maps of CNNs from different dimensions.

To combine advantages of the above methods, switchable normalization (SN)~\cite{C:SN} and its variant~\cite{C:SSN} were proposed to learn linear combination of normalizers for each convolutional layer in an end-to-end manner. We term this normalization setting as static `learning-to-normalize'.
Despite the successes of these methods, once a CNN is optimized by using them, it employed the same combination ratios of the normalization methods for all image samples in a dataset, incapable to adapt to different instances and thus rendering suboptimal performance.

As shown in Fig.~\ref{fig:figure1}, this work studies a new learning problem, that is, dynamic `learning-to-normalize', by proposing Exemplar Normalization (EN), which is able to learn arbitrary normalizer for different convolutional layers, image samples, categories, datasets, and tasks in an end-to-end way.
Unlike previous conditional batch normalization (cBN) that used multi-layer perceptron (MLP) to learn data-dependent parameters in a normalization layer, suffering from over-fitting easily, the internal architecture of EN is carefully designed to learn data-dependent normalization with merely a few parameters, thus stabilizing training and improving generalization capacity of CNNs.

%
%
%
%
%

%
EN has several appealing benefits.
(1) It can be treated as an \textbf{\textit{explanation tool}} for CNNs. The exemplar-based important ratios in each EN layer provide information to analyze the properties of different samples, classes, and datasets in various tasks.
As shown in Fig.~\ref{fig:fig1ratio}, by training ResNet50~\cite{C:Resnet} on ImageNet~\cite{C:ImageNet}, images from different categories would select different normalizers in the same EN layer, leading to superior performance compared to the ordinary network.
%
%
(2)~EN makes \textbf{\textit{versatile design}} of the normalization layer possible, as EN is suitable for various benchmarks and tasks.
Compared with state-of-the-art counterparts in Fig.~\ref{fig:fig1acc},
EN consistently outperforms them on many benchmarks such as  ImageNet~\cite{C:ImageNet} for image classification, Webvision~\cite{A:webvision} for noisy label learning, ADE20K~\cite{C:ADE20K} and Cityscapes~\cite{C:cityscape} for semantic segmentation.
%
(3) EN is a~\textbf{\textit{plug and play module.} } It can be inserted into various CNN architectures such as ResNet~\cite{C:Resnet}, Inception v2~\cite{C:Inception-v3}, and ShuffleNet v2~\cite{C:ShuffleNetV2}, to replace any normalization layer therein and boost their performance.

The  \textbf{contributions} of this work are three-fold.
(1) We present a novel normalization learning setting named dynamic `learning-to-normalize', by proposing Exemplar Normalization (EN), which learns to select different normalizers in different normalization layers for different image samples.
EN is able to normalize image sample in both training and testing stage.
(2) EN provides a flexible way to analyze the selected normalizers in different layers, the relationship among distinct samples and their deep representations.
%
%
%
(3) As a new building block, we apply EN to various tasks and network architectures. Extensive experiments show that EN outperforms its counterparts in wide spectrum of  benchmarks and tasks.
For example, by replacing BN in the ordinary ResNet50~\cite{C:Resnet}, improvement produced by EN is $300\%$ more than that of SN on both ImageNet~\cite{C:ImageNet} and the noisy WebVision~\cite{A:webvision} dataset.


\section{Related  Work}

Many normalization techniques are developed to normalize feature representations~\cite{C:BN,A:LN,A:IN,C:GN,C:SN} or weights of filters~\cite{C:CWN,C:WN,C:SpectralNorm} to accelerate training and boost generation ability of CNNs.
Among them, Batch Normalization (BN)~\cite{C:BN}, Layer Normalization (LN)~\cite{A:LN} and Instance Normalization (IN)~\cite{A:IN} are most popular methods that compute statistics with respect to channel, layer, and minibatch respectively.
%
%
The follow-up Position Normalization~\cite{A:PN} normalizes the activations at each spatial position independently across the channels.
Besides normalizing different dimensions of the feature maps,
another branch of work improved the capability of BN to deal with small batch size, including Group Normalization (GN)~\cite{C:GN}, Batch Renormalization (BRN)~\cite{C:BRN}, Batch Kalman Normalization (BKN)~\cite{C:BKN} and Stream Normalization (StN)~\cite{A:StreamNrom}.

In recent studies, using the hybrid of multiple normalizers in a single normalization layer has achieved much attention~\cite{C:IBN,C:BIN,J:SN,C:SW,C:DN}.
For example, Pan~\textit{et al.} introduced IBN-Net~\cite{C:IBN} to improve the generalization ability of CNNs by manually designing the mixture strategy of IN and BN.
In~\cite{C:BIN}, Nam~\textit{et al.} adopted the same scheme in style transfer, where they employed gated function to learn the important ratios of IN and BN.
Luo~\textit{et al.} further proposed Switchable Normalization (SN)~\cite{C:SN,A:DoNorm} and its sparse version~\cite{C:SSN} to extend such a scheme to deal with arbitrary number of normalizers.
More recently, Dynamic Normalization (DN)~\cite{C:DN} was introduced to estimate the computational pattern of statistics for the specific layer.
Our work is motivated by this series of studies, but provides a more flexible way to learn normalization for each sample.


The adaptive normalization methods are also related to us.
In~\cite{A:CBN}, Conditional Batch Normalization (cBN) was introduced to learn parameters of BN (\ie scale and offset) adaptively as a function of the input features.
%
%
%
Attentive Normalization (AN)~\cite{A:AN} learns sample-based coefficients to combine feature maps.
In~\cite{C:MN}, Deecke~\textit{et al.} proposed Mode Normalization (MN) to detect modes of data on-the-fly and normalize them.
However, these methods are incapable to learn various normalizers for different convolutional layers and images as EN did.
%

The proposed EN also has a connection with learning data-dependent~\cite{C:DynamicFilter} or dynamic weights~\cite{C:L2G} in convolution and pooling~\cite{J:generalizingPooling}.
The subnet for computation of important ratios is also similar to SE-like~\cite{C:SENet,C:AANet,C:ECANet} attention mechanism in form, but they are technically different.
First, SE-like models encourage channels to contribute equally to the feature representation~\cite{A:CENet}, while EN learns to select different normalizers in different layers.
Second, SE is plugged into different networks by using different schemes. EN could directly replace other normalization layers.

%
%
%

\section{Exemplar Normalization (EN)}

\subsection{Notation and Background}

\textbf{Overview.}
We introduce normalization in terms of a 4D tensor, which is the input data of a normalization layer in a mini-batch.
%
Let $\bm{X} \in \mathbb{R}^{N\times C\times H\times W}$ be the input 4D tensor, where $N,C,H,W$ indicate the number of images, number of channels, channel height and width respectively.
Here $H$ and $W$ define the spatial size of a single feature map.
Let matrix $\bm{X}_n \in \mathbb{R}^{C\times HW}$ denote the feature maps of $n$-th image, where $n \in \{1,2,...,N\}$.
Different normalizers normalize $\bm{X}_n$ by removing its mean and standard deviation along different dimensions, performing a formulation
\begin{equation}
\widehat{\bm{X}}_n = \bm{\gamma} ~ \frac{ \bm{X}_n - \bm{\mu}^k }{ \sqrt{(\bm{\delta}^k)^2 + \epsilon } } + \bm{\beta}
\end{equation}
where $\widehat{\bm{X}}_n$ is the feature maps after normalization.
$\bm{\mu}^k$ and $\bm{\delta}^k$ are the vectors of mean and standard deviation calculated by the $k$-th normalizer.
%
%
%
Here we define $k\in\{$BN, IN, LN, GN,...$\}$.
The scale parameter $\bm{\gamma} \in \mathbb{R}^C$ and bias parameter $\bm{\beta} \in \mathbb{R}^C$ are adopted to re-scale and re-shift the normalized feature maps.
$\epsilon$ is a small constant to prevent dividing by zero,
and both $\sqrt{\cdot}$ and $(\cdot)^2$ are channel-wise operators.

\textbf{Switchable Normalization (SN).}
Unlike previous methods that estimated statistics over different dimensions of the input tensor, SN~\cite{C:SN,J:SN} learns a linear combination of statistics of existing normalizers,
\begin{equation}
\widehat{\bm{X}}_n = \bm{\gamma} ~ \frac{ \bm{X}_n - \sum_k \lambda^k \bm{\mu}^k }{ \sqrt{  \sum_k \lambda^k ~   (\bm{\delta}^k)^2 + \epsilon } } + \bm{\beta}
\label{eq:sn}
\end{equation}
where $\lambda^k \in [0,1]$ is a learnable parameter corresponding to the $k$-th normalizer, and $\sum_k \lambda^k = 1$.
In practice, this important ratio is calculated by using the softmax function.
The important ratios for mean and variance can  be also different.
Although SN~\cite{C:SN} outperforms the individual normalizer in various tasks,
it solves a static `learning-to-normalize' problem by switching among several normalizers in each layer.
Once SN is learned, its important ratios are fixed for the entire dataset.
Thus the flexibility of SN is limited and it suffers from the bias between the training and the test set, leading to sub-optimal results.
%

In this paper, Exemplar Normalization (EN) is proposed to investigate a dynamic `learning-to-normalize' problem, which learns different data-dependant normalizations for different image samples in each layer.
%
EN extremely expands the flexibility of SN, while retaining SN's advantages of differential learning, stability of model training, and capability in multiple tasks.

\subsection{Formulation of EN}

Given input feature maps $\bm{X}_n$,  Exemplar Normalization (EN) is defined by

\begin{equation}
\widehat{\bm{X}}_n = \sum_k  ~~\bm{\gamma}^k (~\lambda^k_n~ \frac{ \bm{X}_n - \bm{\mu}^k }{ \sqrt{  (\bm{\delta}^k)^2 + \epsilon } }~)+ \bm{\beta}^k
\label{eq:csn}
\end{equation}
where $\lambda^k_n \in [0,1]$ indicates the important ratio of the $k$-th normalizer for the $n$-th sample.
Similar with SN, we use softmax function to satisfy the summation constraint, $\sum_k \lambda^k_n = 1$.
Compared with Eqn.~\eqref{eq:sn} and Eqn.~\eqref{eq:csn}, the differences between SN and EN are three-fold.
(1) The important ratios of mean and standard deviation in SN can be different, but such scheme is avoided in EN to ensure stability of training, because the learning capacity of EN already outperforms SN by learning different normalizers for different samples.
%
%
(2) We use important ratios to combine the normalized feature maps instead of combining statistics of normalizers, reducing the bias in SN when combining the standard deviations.
(3) Multiple $\bm{\gamma}$ and $\bm{\beta}$ are adopted to re-scale and re-shift the normalized feature maps in EN.

To calculate the important ratios $\lambda^k_n$ depended on the feature map of individual sample,
we define
\begin{equation}
\bm\lambda_n = \mathcal{F}( \bm{X}_n, \bm{\Omega}; \Theta)
\end{equation}
where $\bm\lambda_n = [\lambda^1_n,..., \lambda^k_n,...\lambda^K_n]$, and $K$ is the total number of normalizers in EN.
$\bm{\Omega}$ indicates a collection of statistics of different normalizers. We have $\bm{\Omega}=\{ (\bm{\mu}^k, \bm{\delta}^k)\}_{k=1}^K$.
$\mathcal{F}(\cdot)$ is a function (a small neural network) to calculate the instance-based important ratios, according to the input feature maps $\bm{X}_n$ and statistics $\bm{\Omega}$.
$\Theta$ denotes learnable parameters of function $\mathcal{F}(\cdot)$.
We carefully design a lightweight module to implement the function $\mathcal{F}(\cdot)$ in next subsection.
%

\begin{figure}[t]
\begin{center}
\includegraphics[width=0.65\linewidth]{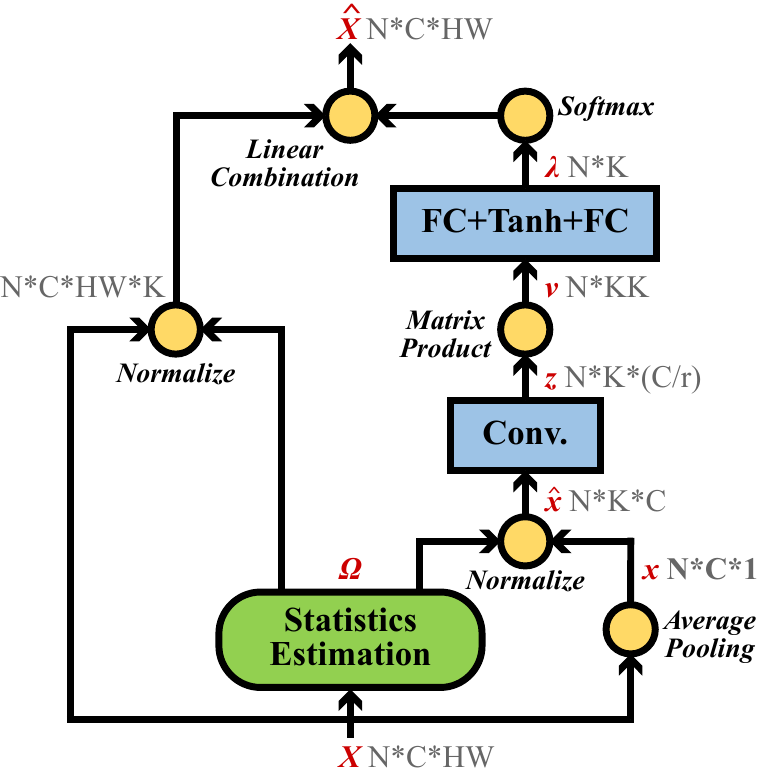}
\end{center}
\caption{Illustration of the Exemplar Normalization (EN) layer, which is able to learn the sample-based important ratios to normalize the input feature maps by using multiple normalizers.
Note that the scale parameter $\bm{\gamma}$ and shift parameter $\bm{\beta}$ in Eqn.~\eqref{eq:csn} are omitted to simplify the diagram.}
\label{fig:CSN}
\vspace{-0.4cm}
\end{figure}

\subsection{An Exemplar Normalization Layer}
\label{sec:csnlayer}

Fig.~\ref{fig:CSN} shows a diagram of the key operations in an EN layer, including important ratio calculation and feature map normalization.
Given an input tensor $\bm{X}$, a set of statistics $\bm{\Omega}$ are estimated.
We use $\bm{\Omega}_k$ to denote the $k$-th statistics (mean and standard deviation).
Then the EN layer uses $\bm{X}$ and $\bm{\Omega}$ to calculate the important ratios  as shown in the right branch of Fig.~\ref{fig:CSN} in blue. As shown in the left branch of Fig.~\ref{fig:CSN}, multiple normalized tensors are also calculated.
%
%

In Fig.~\ref{fig:CSN}, there are three steps to calculate the important ratios for each sample.
(1) The input tensor $\bm{X}$ is firstly down-sampled in the spatial dimension by using average pooling.
The output feature matrix is denoted as $\bm{x} \in \mathbb{R}^{N\times C} $.
%
%
Then we use every $\bm{\Omega}_k$ to pre-normalize $\bm{x}$  by subtracting the means and dividing by the standard deviations.
There are $K$ statistics and thus we have $\hat{\bm{x}} \in \mathbb{R}^{N \times K \times C}$.
After that, a 1-D convolutional operator is employed to reduce the channel dimension of  $\hat{\bm{x}}$ from $C$ to $C/r$, which is shown in the first blue block in Fig.~\ref{fig:CSN}.
Here $r$ is a hyper-parameter that indicates the reduction rate.
%
%
To further reduce the parameters in the above operation, we use group convolution with the group number $C/r$ to ensure the total number of convolutional parameters always equals to $C$, irrelevant to the value of $r$.
The output in this step is denoted as $\bm{z}$.
%
%
%
%
%

(2) The second step is to compute the pairwise correlation of different normalizers for each sample, which is motivated by the high-order feature representation~\cite{C:HighOrder,C:Anet}.
For the $n$-th sample, we use $\bm{z}_n \in \mathbb{R}^{K\times C}$ and its transposition $\bm{z}_n^T$ to compute the pairwise correlations by $\bm{v}_n = \bm{z}_n \bm{z}_n^T\in \mathbb{R}^{K\times K}$.
%
%
%
Then $\bm{v}_n$ is reshaped to a vector to calculate the important ratios.
%
%
Intuitively, the pairwise correlations capture the relationship between different normalizers for each sample,
and allow the model to integrate more information to calculate the important ratios.
In practice, we also find such operation could effectively stabilize the model training and make the model achieve higher performance.

(3) In the last step, the above vector $\bm{v}_n$ is firstly fed into a fully-connected (FC) layer followed by a tanh unit. This is to raise its dimensions to $\pi K$,
%
where $\pi$ is a hyper-parameter and the value of $K$ is usually small, \eg $3$.
In practice, we set the value of $\pi$ as $50$ in experiments.
After that, we perform another FC layer to reduce the dimension to $K$.
The output vector $\bm{\lambda}_n\in\mathbb{R}^{K\times 1 }$ is regarded as the important ratios of the $n$-th sample for $K$ normalizers,
where each element is corresponding to an individual normalizer.
Once we obtain the important ratio $[\bm{\lambda}_1,\bm{\lambda}_2,...,\bm{\lambda}_N]^T $, the \texttt{softmax} function is applied to satisfy the summation constraint that the important ratios of different normalizers sum to $1$.
%
%


\textbf{Complexity Analysis.}
The numbers of parameters and computational complexity of different normalization methods are compared in Table~\ref{tab:complexity}.
The additional parameters in EN are mainly from the convolutional and FC layers to calculate the data-dependant important ratios.
In SN~\cite{C:SN}, such number is $2K$ since it adopts the global important ratios for both mean and standard deviant.
In EN, the total number of parameters that is applied to generate the data-dependant important ratios is $C+\Psi(K)$,
where $C$ equals to the input channel size of the convolutional layer (\ie ``Conv.'' with $C$ parameters  in Fig.~\ref{fig:CSN}).
$\Psi(K)$ is a function of $K$, which indicates the amount of parameters in the two FC layers (\ie the top blue block in Fig.~\ref{fig:CSN}).
In practice, since the number of $K$ is  small (\eg $3\sim4$), the value of $\Psi(K)$ is just about $0.001M$.
In this paper, EN employ a pool of normalizers that is the same as SN, \ie $\{$IN,LN,BN$\}$.
Thus the computational complexities of both SN and EN for estimating the statistics are $\mathcal{O}(NCHW)$.
%
We also compare FLOPs in Sec.~\ref{sec:Experiment}, showing that the extra \#parameters of EN is marginal compared to SN, but its relative improvement over the ordinary BN is 300\% larger than SN.

\begin{table}[t]
\small
\begin{center}
\caption{\textbf{Comparisons} of parameters and computational complexity of different normalizers.
$\bm{\gamma}$ and $\bm{\beta}$ indicate the scale and shift parameters in Eqn.\eqref{eq:sn}, and $\bm{\Theta}$ is the parameters of ``Conv.'' and FC layer in proposed EN.
$K$ denotes the number of normalizer and $\Psi(\cdot)$ is a function of $K$ that determines the number of $\bm{\Theta}$.
$\{\omega_k, \nu_k\}_{k=1}^K$ are the learnable important ratios in SN~\cite{C:SN}. }
\begin{tabular}{c|cp{1.5cm}p{1.5cm}}
\hline
\multirow{2}{*}{Method} & \multirow{2}{*}{params} &  \multirow{2}{*}{$\#$params}  & computation   \\
& & &complexity \\
\hline
BN~\cite{C:BN} & $\bm{\gamma},\bm{\beta}$ & $2C$  & $\mathcal{O}(NCHW)$\\
IN~\cite{A:IN} & $\bm{\gamma},\bm{\beta}$ & $2C$  & $\mathcal{O}(NCHW)$ \\
LN~\cite{A:LN} & $\bm{\gamma},\bm{\beta}$ & $2C$  & $\mathcal{O}(NCHW)$ \\
GN~\cite{C:GN} & $\bm{\gamma},\bm{\beta}$ & $2C$  & $\mathcal{O}(NCHW)$ \\
BKN~\cite{C:BKN} &  $\bm{A}$ & $C^2$ & $\mathcal{O}(NC^2HW)$ \\
SN~\cite{C:SN} & $\bm{\gamma},\bm{\beta}, \{\omega_k, \nu_k\}_{k=1}^K $ & $2C + 2K$  & $\mathcal{O}(NCHW)$ \\
\hline
\multirow{2}{*}{EN} &  \multirow{2}{*}{$\bm{\gamma},\bm{\beta},\bm{\Theta}$} &$2KC+   $  &  \multirow{2}{*}{$\mathcal{O}(NCHW)$ } \\
                    &  & $ C+ \Psi(K) $ & \\
\hline
\end{tabular}
\label{tab:complexity}
\end{center}
\vspace{-0.2cm}
\end{table}

\section{Experiment}
\label{sec:Experiment}

%

\subsection{Image Classification with ImageNet dataset}
\label{sec:imagenet}



\noindent
\textbf{Experiment Setting.}
We first examine the performance of proposed EN on ImageNet~\cite{C:ImageNet}, a standard large-scale dataset for high-resolution image classification.
Following~\cite{C:SN}, the $\bm{\gamma}$ and $\bm{\beta}$ in all of the normalization methods are initialized as $1$ and $0$ respectively.
In the training phase, the batch size is set as $128$ and the data augmentation scheme is employed same as ~\cite{C:Resnet} for all of the methods.
In inference, the single-crop validation accuracies based on $224\times224$ center crop are reported.

We use ShuffleNet v2 x$0.5$~\cite{C:ShuffleNetV2} and ResNet50~\cite{C:Resnet} as the backbone network to evaluate various normalization methods since the difference in their network architectures and the number of parameters.
Same as~\cite{C:ShuffleNetV2}, ShuffleNet v2 is trained by using Adam optimizer with the initial learning rate $0.1$.
For ResNet50, all of the methods are optimized by using stochastic gradient decent (SGD) with stepwise learning rate decay.
The hyper-parameter $r$ in ShuffleNet v2 x$0.5$ and ResNet50 are set as $8$ and $32$ respectively since the smallest number of channels are different.
The hyper-parameter $\pi$ is $50$.
For fair comparison, we replace compared normalizers with EN in all of the normalization layers in the backbone network.
%

%


\begin{table}[t]
\small
\begin{center}
\caption{ Comparisons of classification accuracies ($\%$), network parameters (Params.) and floating point operations per second (GFLOPs) of various methods on the validation set of ImageNet by using different network architectures. }  
\begin{tabular}{c|c|c c c c }
\hline
Backbone& Method & GFLOPs & Params. & top-1 & top-5 \\
\hline\hline
              & BN& 0.046 & 1.37M &60.3 & 81.9\\
ShuffleNet  & SN& 0.057 & 1.37M &61.2 & 82.9\\
v2 x0.5          &SSN& 0.052 & 1.37M &61.2 & 82.7\\
              &EN & 0.063 & 1.59M &\textbf{62.2}& \textbf{83.3}\\
\hline
\hline
         & SENet  & 4.151 &  26.77M & 77.6 &  93.7   \\
         & AANet  & 4.167 &  25.80M & 77.7 & \textbf{93.8} \\
\cline{2-6}
          & BN& 4.136 & 25.56M & 76.4 & 93.0 \\
          & GN& 4.155 & 25.56M & 76.0 & 92.8 \\
ResNet50  & SN& 4.225 & 25.56M & 76.9 & 93.2 \\
          &SSN& 4.186 & 25.56M & 77.2 & 93.1\\
          &EN & 4.325 & 25.91M & \textbf{78.1}& 93.6 \\
\hline
\end{tabular}
\vspace{-0.2cm}
\label{table:imagenet}
\end{center}
\end{table}

\begin{figure}[htbp]
\centering
\subfigure[ResNet20 on CIFAR-10]{
\begin{minipage}[t]{0.5\linewidth}
\centering
\includegraphics[width=0.85\linewidth]{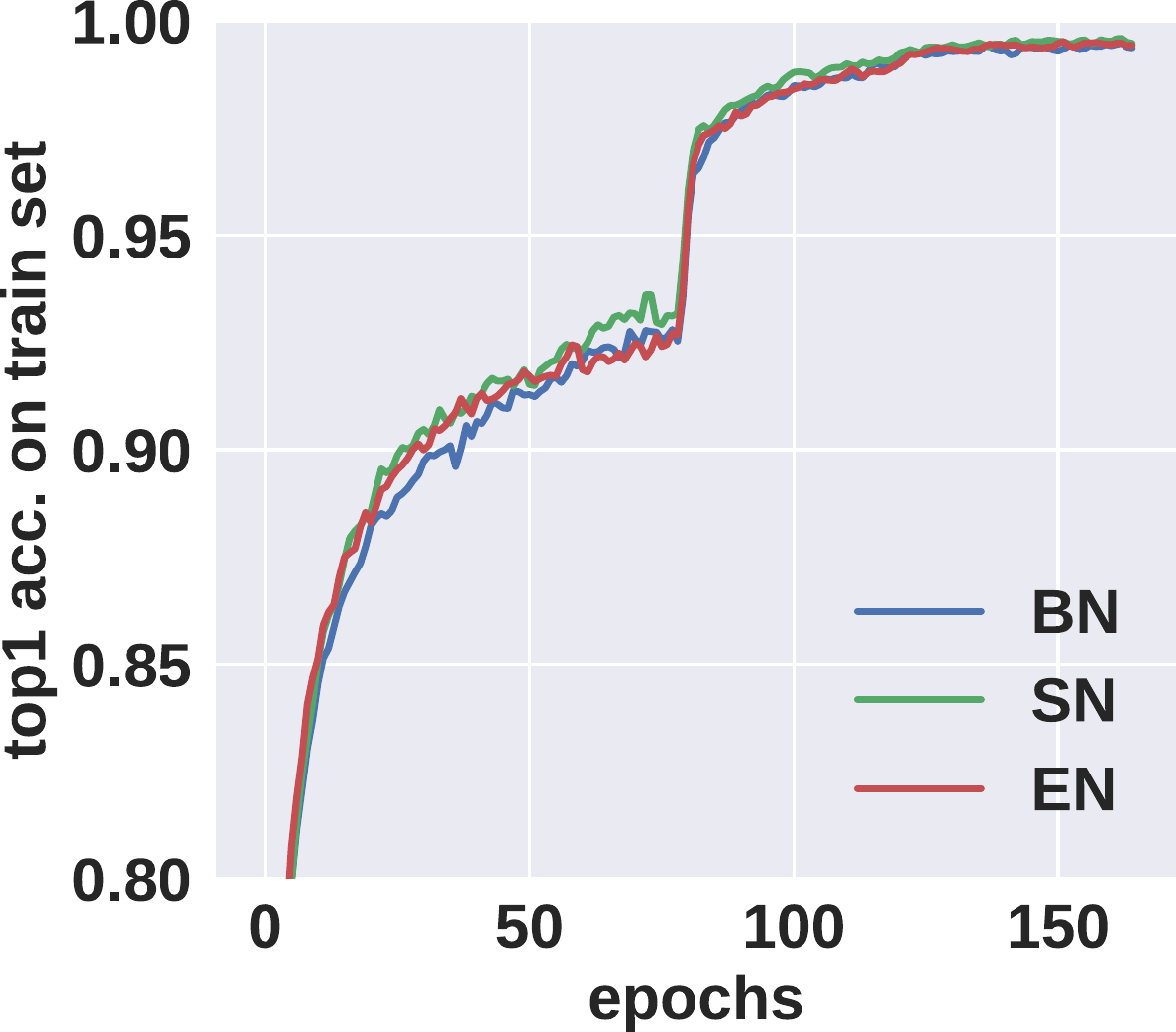}
\end{minipage}%
}%
\subfigure[ResNet20 on CIFAR-10]{
\begin{minipage}[t]{0.5\linewidth}
\centering
\includegraphics[width=0.85\linewidth]{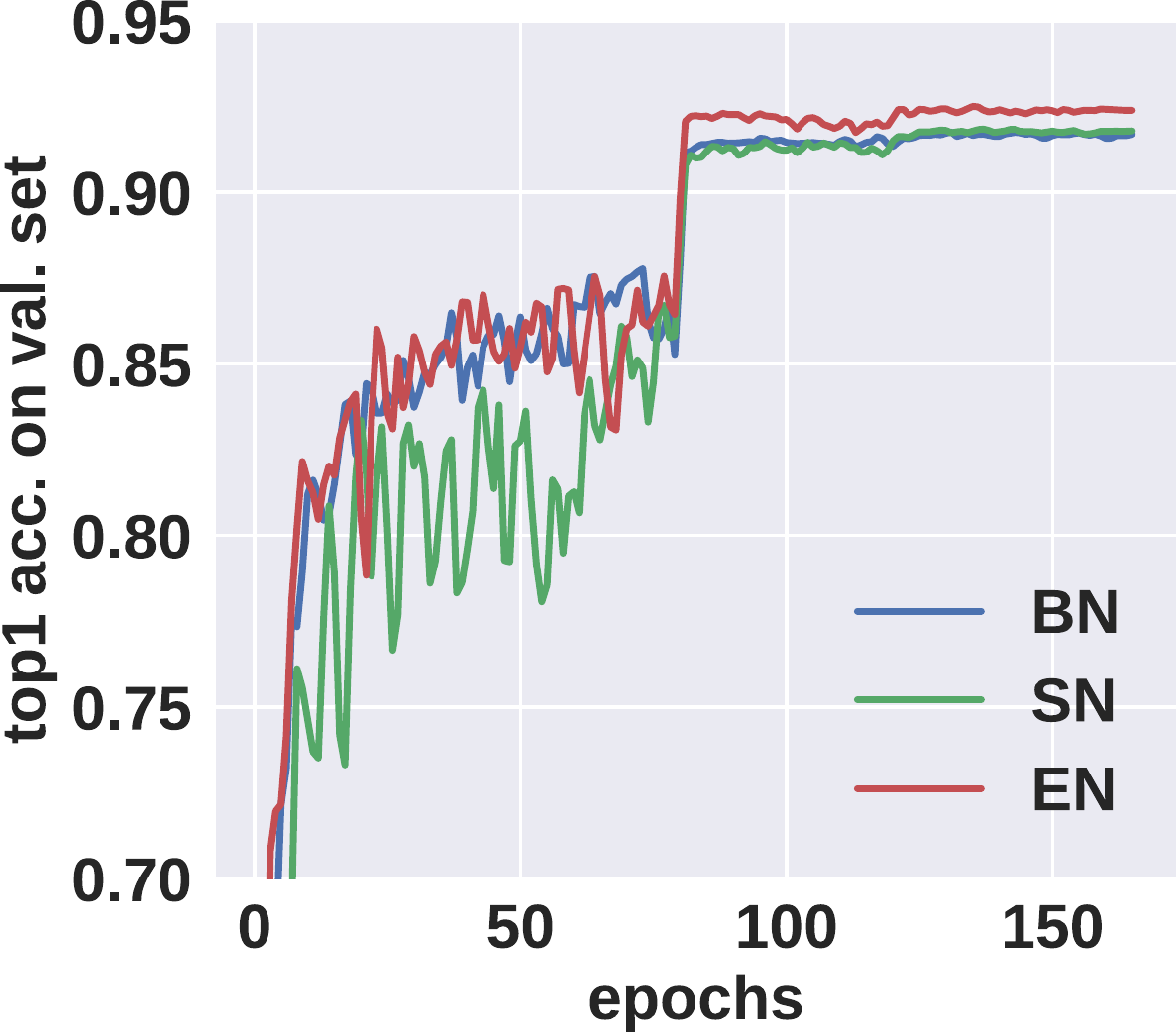}
\end{minipage}%
}   \\
\subfigure[ResNet50 on ImageNet]{
\begin{minipage}[t]{0.5\linewidth}
\centering
\includegraphics[width=0.9\linewidth]{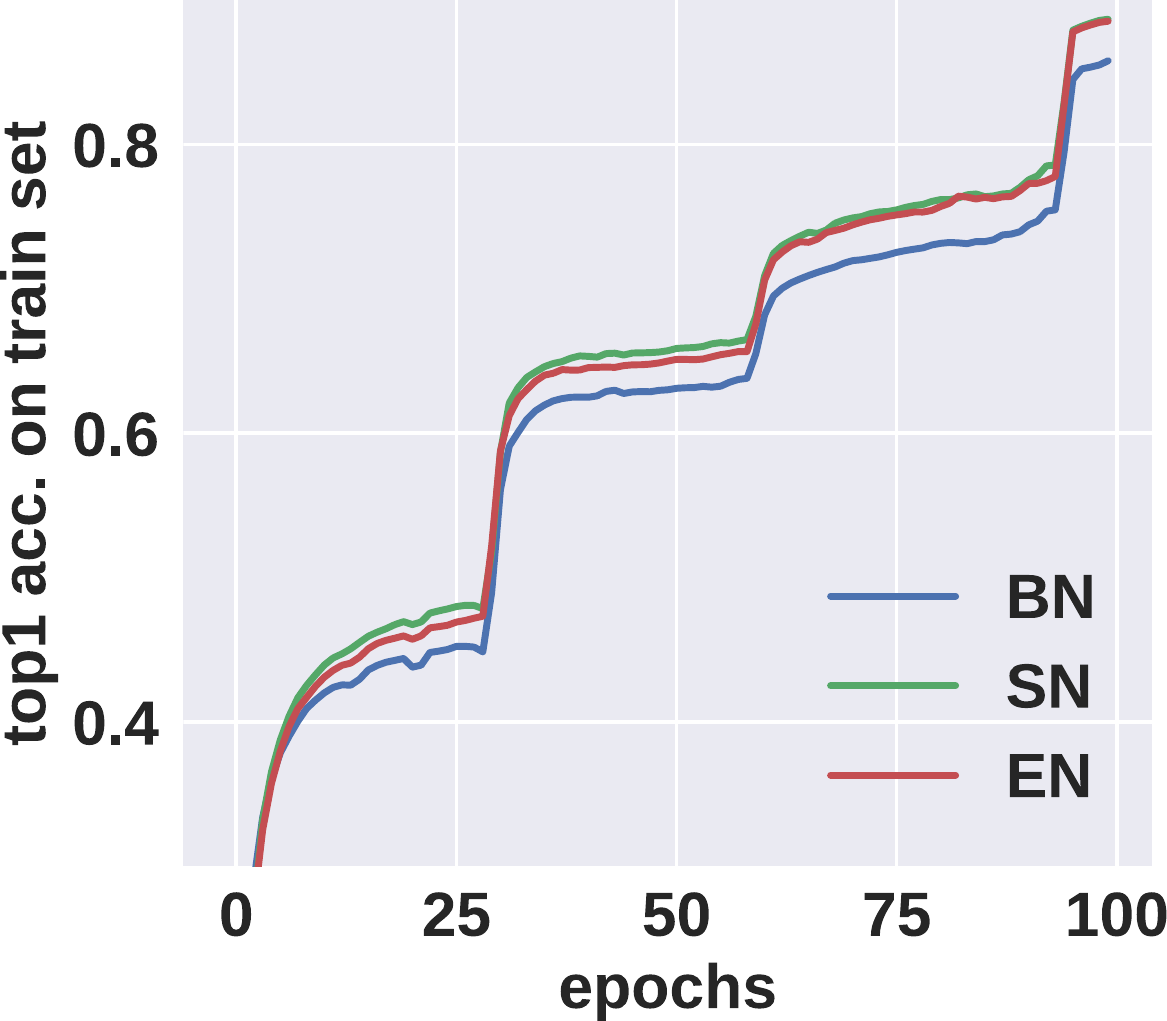}
\end{minipage}
}%
\subfigure[ResNet50 on ImageNet]{
\begin{minipage}[t]{0.5\linewidth}
\centering
\includegraphics[width=0.85\linewidth]{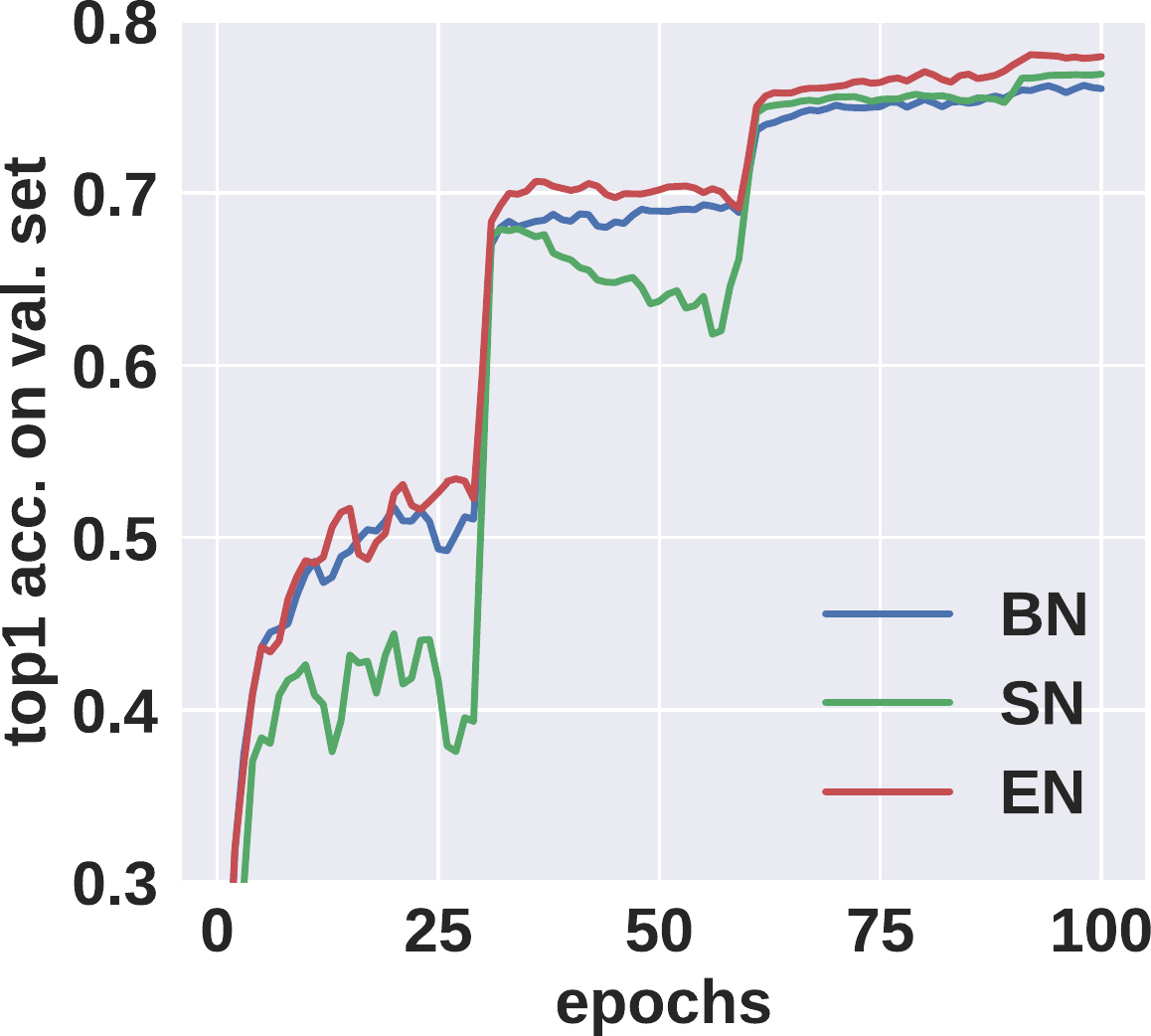}
\end{minipage}
} \\
\subfigure[ResNet50 on Webvision]{
\begin{minipage}[t]{0.5\linewidth}
\centering
\includegraphics[width=0.85\linewidth]{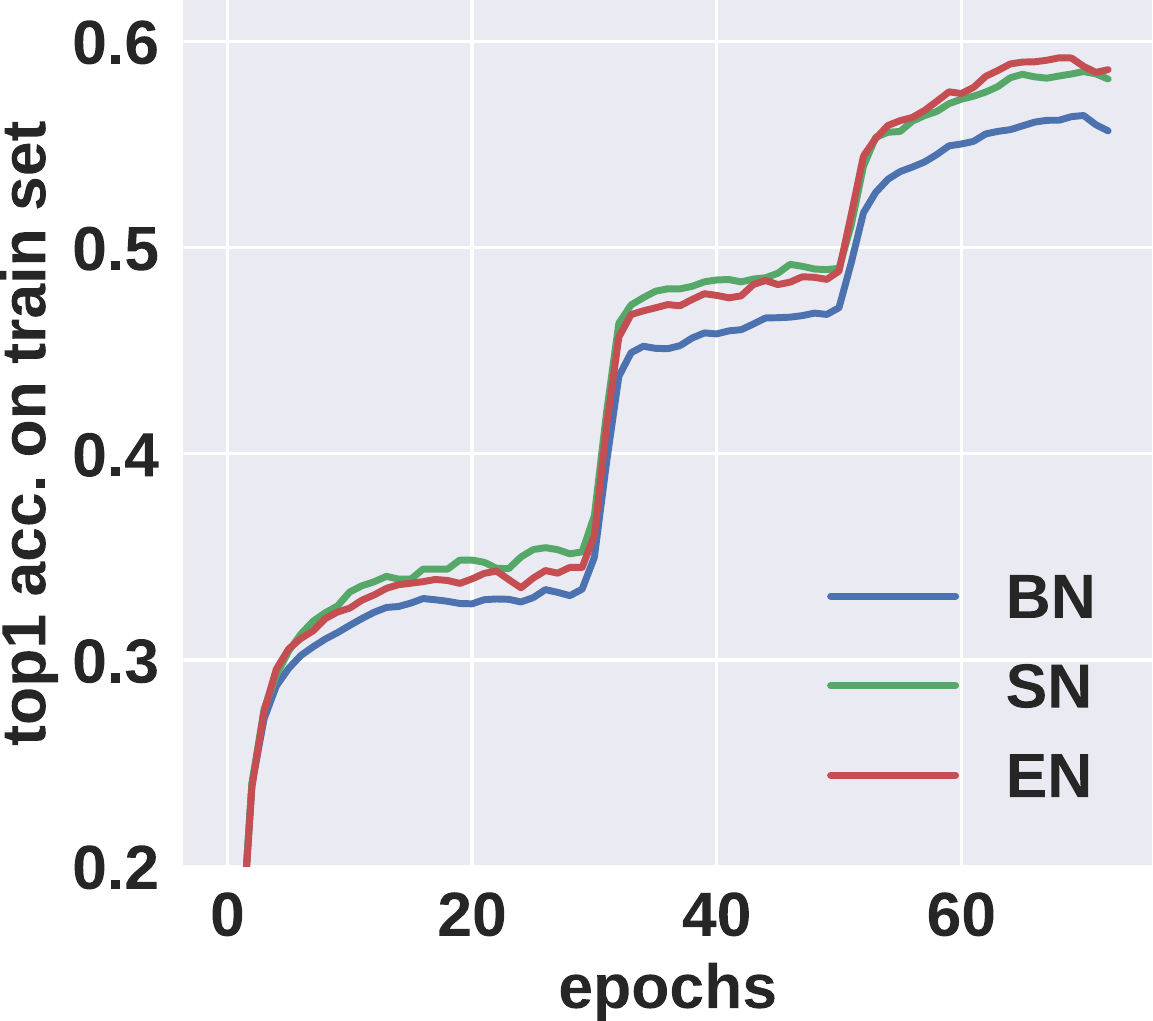}
\end{minipage}
}%
\subfigure[ResNet50 on Webvision]{
\begin{minipage}[t]{0.5\linewidth}
\centering
\includegraphics[width=0.85\linewidth]{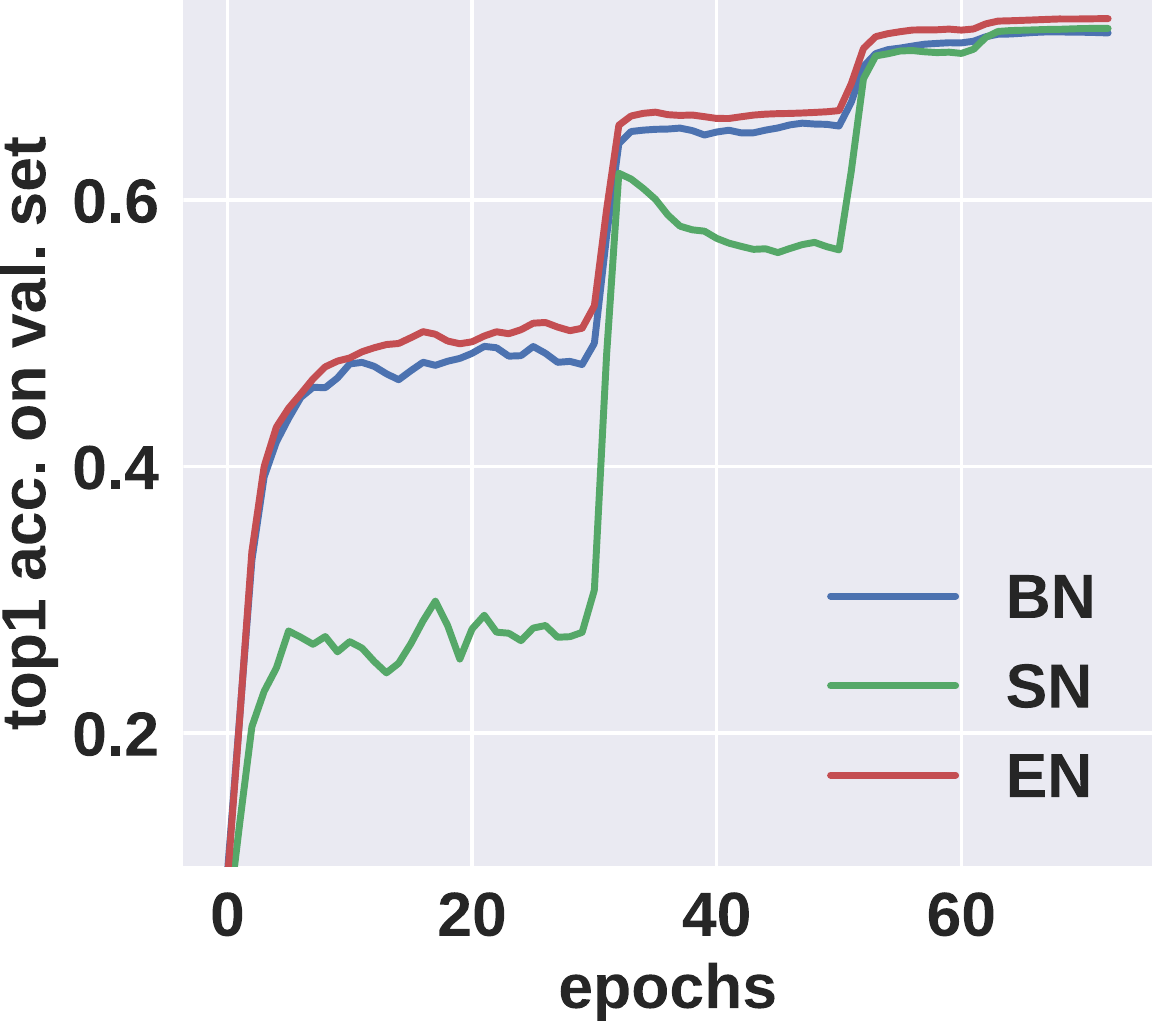}
\end{minipage}
}
\centering
\vspace{-0.3cm}
\caption{Top-1 training and validation accuracy curves of different normalization methods on CIFAR-10, ImageNet and Webvision dataset. Zoom in three times for the best view.}
\label{fig:threeDatasetCurve}
\end{figure}

\noindent
\textbf{Result Comparison.}
Table~\ref{table:imagenet} reports the efficiency and accuracy of EN against its counterparts including BN~\cite{C:BN}, GN~\cite{C:GN}, SN~\cite{C:SN} and SSN~\cite{C:SSN}.
For both two backbone networks, EN offers a super-performance and a competitive computational cost compared with previous methods.
For example, by considering the sample-based ratio selection, EN outperforms SN $1.0\%$, and $1.2\%$ on top-1 accuracy by using ShuffleNet v2 x0.5 and ResNet50 with only a small amount of GFLOPs increment.
The top-1 accuracy curves of ResNet50 by using BN, SN and EN on training and validation set of ImageNet are presented in Fig.~\ref{fig:threeDatasetCurve}.
We also compare the performance with state-of-the-art attention-based methods, \ie SENet~\cite{C:SENet} and AANet~\cite{C:AANet}, without bells and whistles, the proposed EN still outperforms these methods.

\subsection{Noisy Classification with Webvision dataset}
\label{sec:webvision}

\noindent
\textbf{Experiment Setting.}
We also evaluate the performance of EN on noisy image classification task with Webvision dataset~\cite{A:webvision}.
We adopt Inception v2~\cite{C:Inception-v3} and ResNet50~\cite{C:Resnet} as the backbone network.
Since the smallest number of channels in Inception v2 is $32$, the feature reduction rate $r$ in the first ``Conv.'' is set as $16$ for such network architecture.
In ResNet50~\cite{C:Resnet}, we maintain the same reduction parameter $r=32$ as Imagenet.
The center crop with the image size $224 \times 224$ are adopted in inference.
All of the models are optimized with SGD, where the learning rate is initialized as $0.1$ and decreases at the iterations of $\{30,50,60,65,70\}\times10^4$ with a factor of $10$.
The batch size is set as $256$ and the data augmentation and data balance technologies are used by following~\cite{C:CurriculumNet}.
In the training phase, we replace compared normalizers with EN in all of the normalization layers.


\noindent
\textbf{Result Comparison.}
Table~\ref{table:webvision} reports the top-1 and top-5 classification accuracies of various normalization methods.
EN outperforms its counterparts by using both of two network architectures.
Specially, by using ResNet50 as the backbone, EN significantly boost the top-1 accuracy from $72.8\%$ to $73.5\%$ compared with SN.
It achieves about $3$ times  relative improvement of EN against SN compared to the ordinary plain ResNet50.
Such performance gain is consistent with the results on ImageNet.
The training and validation curves are shown in Fig.~\ref{fig:threeDatasetCurve}.

The cross dataset test is also conducted to investigate the transfer ability of EN since the categories in ImageNet and Webvision are the same.
The model trained on one dataset is used to do the test on another dataset's validation set.
The results are reported in Fig.~\ref{tab:crossdataset} that EN still outperforms its counterparts.

\begin{table}[t]
\small
\begin{center}
\caption{ Comparison of classification accuracies ($\%$), network parameters and GFLOPs of various normalization methods on the validation set of Webvision by using different network architectures. The best results are bold. }
\begin{tabular}{c|c|c c c c }
\hline
Model& Norm & GFLOPs & Params. & top-1 & top-5 \\
\hline\hline
\multirow{3}{*}{ Inception v2}
& BN& 2.056 & 11.29M & 70.7 & 88.0 \\
& SN& 2.081 & 11.30M& 71.3 & 88.5 \\
& EN& 2.122 & 12.36M& \textbf{71.6 }& \textbf{88.6}\\
\hline
\multirow{3}{*}{ ResNet50}
& BN& 4.136 & 25.56M  &72.5 & 89.1\\
& SN& 4.225 & 25.56M  &72.8 & 89.2\\
& EN& 4.325 & 25.91M  &\textbf{73.5} & \textbf{89.4}\\
\hline
\end{tabular}
\label{table:webvision}
\end{center}
\vspace{-0.3cm}
\end{table}

\begin{table}[t]
\small
\begin{center}
\caption{Top-1 and top-5 accuracy ($\%$) of cross dataset results. The dataset before '$\rightarrow$' is adopted to train ResNet50 with various normalization methods. The validation set after '$\rightarrow$' is used for testing. The number of categories in two datasets are the same.}
\begin{tabular}{c|c|cc}
\hline
training set $\rightarrow$ val. set & method   &  top-1 & top-5 \\
\hline
\hline
 \multirow{3}{*}{ImageNet$\rightarrow$ Webvision }  &   BN       &  67.9 & 85.8  \\
                                                    &   SN       &  68.0  & 86.3  \\
                                                   &    EN       &  \textbf{68.4}  & \textbf{86.8}  \\
\hline
 \multirow{3}{*}{Webvision $\rightarrow$ ImageNet}  &   BN       &  64.4  & 84.3  \\
                                                   &    SN       &  61.1  & 81.0  \\
                                                    &   EN       &  \textbf{64.7}  & \textbf{84.6}  \\
\hline
\end{tabular}
\label{tab:crossdataset}
\end{center}
\vspace{-0.3cm}
\end{table}

\subsection{Tiny Image Classification with CIFAR dataset}
\label{sec:cifar}

\noindent
\textbf{Experiment Setting.}
We also conduct the experiment on CIFAR-10 and CIFAR-100 dataset.
The training batch size is $128$.
All of the models are trained by using the single GPU.
The training process contains $165$ epoches.
The initial learning rate is set as $0.1$ and decayed at $80$ and $120$ epoch, respectively.
We also adopt the warm up scheme~\cite{C:Resnet,C:ResV2} for all of the models training,
which increases the learning rate from $0$ to $0.1$ in the first epoch.

\noindent
\textbf{Result Comparison.}
The experiment results on CIFAR dataset are presented in Table~\ref{tab:cifar}.
Compared with the previous methods, EN shows better performance than the other normalization methods over various depths of ResNet~\cite{C:Resnet}.
In particular, the top-1 accuracies of EN on CIFAR-100 are significantly improved by $1.04\%$, $1.31\%$ and $0.79\%$ compared with SN with different network depths.

\begin{table}[t]
\small
\begin{center}
\caption{Top-1 accuracy ($\%$) on CIFAR-10  and  CIFAR-100 dataset by using various networks. The best results are bold.}
\begin{tabular}{c|c|ccc}
\hline
Dataset& Backbone & BN  & SN & EN\\
\hline
\hline
\multirow{3}{*}{ CIFAR-10}
& ResNet20  & 91.54 & 91.81 & \textbf{92.41}\\
& ResNet56  & 93.15  & 93.41 &\textbf{93.73} \\
& ResNet110 & 93.88& 94.01 &  \textbf{94.22}\\
\hline
\multirow{3}{*}{ CIFAR-100}
& ResNet20 & 67.87 &67.74 & \textbf{68.78} \\
& ResNet56& 70.83 & 70.70  & \textbf{72.01}\\
& ResNet110& 72.41 & 72.53 & \textbf{73.32}\\
\hline
\end{tabular}
\label{tab:cifar}
\end{center}
\vspace{-0.3cm}
\end{table}

\subsection{Semantic Image Segmentation}

\noindent
\textbf{Experiment Setting.}
We also evaluate the performance of EN on semantic segmentation task by using standard benchmarks, \ie ADE20K~\cite{C:ADE20K} and Cityscapes~\cite{C:cityscape} datasets, to demonstrate its generalization ability.
Same as~\cite{C:SN,J:PDN}, we use DeepLab~\cite{J:deeplab} with ResNet50 as the backbone network and adopt the atrous convolution with the rate $2$ and $4$ in the last two blocks.
The downsample rate of the backbone network is $8$ and the bilinear operation is employed to upsample the predicted semantic maps to the size of the input image.
All of the models are trained with $2$ samples per GPU by using ``ploy'' learning rate decay.
The initial learning rate on ADE20K and Cityscapes are set as $0.02$ and $0.01$, respectively.
Single-scale and multi-scale testing are used for evaluation.
Note that the synchronization scheme is not used in SN and EN to estimate the batch mean and batch standard deviate across multi-GPU.
To finetune the model on semantic segmentation, we use $8$ GPU with $32$ images per GPU to pre-train the EN-ResNet50 in ImageNet, thus we report the same configuration of SN (\ie SN(8,32)~\cite{J:SN}) for fair comparision.

\noindent
\textbf{Result Comparison.}
The mIoU scores on ADE20K validation set and Cityscapes test set are reported in Table~\ref{tab:segmenation}.
The performance improvement of EN is consistent with the results in classification.
For example, the mIoUs on ADE20K and Cityscapes are improved from $38.4\%$ and $75.8\%$ to $38.9\%$ and $76.1\%$ by using multi-scale test.

\begin{table}[t]
\small
\begin{center}
\caption{Semantic Segmentation results on  ADE20K and Cityscapes datasets. The backbone is ResNet50 with dilated convolutions. The subscripts ``ss" and ``ms" indicate single-scale and multi-scale test respectively. The best results are bold.}
\begin{tabular}{l|cc|cc}
\hline
\multirow{2}{*}{ Method}
& \multicolumn{2}{c|}{ ADE20K } &\multicolumn{2}{c}{ Cityscapes } \\
\cline{2-5}
& mIoU$_{ss}$ & mIoU$_{ms}$ & mIoU$_{ss}$ &  mIoU$_{ms}$\\
\hline
\hline
SyncBN & 36.4 & 37.7 & 69.7 & 73.0 \\
GN     & 35.7 & 36.6 & 68.4 & 73.1 \\
SN     & 37.7 & 38.4 & 72.2 & 75.8 \\
EN     & \textbf{38.2} & \textbf{38.9} & \textbf{72.6} & \textbf{76.1} \\
\hline
\end{tabular}
\label{tab:segmenation}
\end{center}
\vspace{-0.3cm}
\end{table}

\subsection{Ablation Study}

\textbf{Hyper-parameter $\pi$}.
We first investigate the effect of hyper-parameter $\pi$ in Sec.~\ref{sec:csnlayer}.
The top-1 accuracy on ImageNet by using ResNet50 as the backbone network are reported in Table~\ref{tab:hyperpara1}.
All of the EN models outperform SN.
With the number of $\pi$ increasing, the performance of classification growths steadily.
The the gap between the lowest and highest is about $0.6\%$ excluding $\pi=1$, which demonstrates the model is not sensitive to the hyper-parameter $\pi$ in most situations.
To leverage the classification accuracy and computational efficiency, we set $\pi$ as $50$ in our model.

\begin{table}[t]
\small
\begin{center}
\caption{Top-1  accuracy ($\%$)  on  ImageNet  by  using  EN-ResNet50 with different ascending dimension hyper-parameter $\pi$. }
\begin{tabular}{c|c|ccccc}
\hline
\multirow{2}{*}{Method}&\multirow{2}{*}{ SN} & \multicolumn{5}{c}{ EN ( value of hyper-parameter $\pi$ ) }  \\
\cline{3-7}
             &   & 1 & 10 & 20 &  50 & 100 \\
\hline
top-1            & 76.9 & 77.1    & 77.5   & 77.8   & \textbf{78.1}    & 78.0 \\
$\Delta$ \vs SN  & -    & $+$ 0.2 & $+$ 0.6& $+$ 0.9 & $\textbf{+1.2}$ & $+$ 1.1\\
\hline
\end{tabular}
\label{tab:hyperpara1}
\end{center}
\vspace{-0.3cm}
\end{table}

\begin{table}[t]
\small
\begin{center}
\caption{Top-1 accuracy ($\%$) on ImageNet by using EN-ResNet50 with different hyper-parameter $r$ in the `Conv.' of Sec.~\ref{sec:csnlayer}. Note that the total number of parameters with different $r$ are the same. }
\begin{tabular}{c|c|ccccc}
\hline
\multirow{2}{*}{Method}&\multirow{2}{*}{ SN} & \multicolumn{5}{c}{ EN ( value of hyper-parameter $r$ ) }  \\
\cline{3-7}
& & 2 & 4 & 16 &  32 & 64 \\
\hline
top-1            & 76.9 & 77.7  & 77.9  &77.9   & \textbf{78.1}   & 77.7 \\
$\Delta$ \vs SN  & -    & $+0.8$& $+1.0$& $+1.0$& $\textbf{+1.2}$ & $+0.8$ \\
\hline
\end{tabular}
\label{tab:hyperpara2}
\end{center}
\vspace{-0.3cm}
\end{table}

\begin{table}[t]
\small
\begin{center}
\caption{Top-1 and Top-5 accuracy ($\%$) on ImageNet by using EN-ResNet50 with different configurations.}
\begin{tabular}{p{2.4cm}|cc}
\hline
\multirow{2}{*}{Method} & \multirow{2}{*}{top-1~/~top5 } & top-1~/~top5       \\
                        &                                & $\Delta$ \vs EN    \\
\hline
\hline
EN-ResNet50         & 78.1~/~93.6 & - \\
\hline
$a.$ $\rightarrow$ 2-layer MLP                & 76.7~/~92.9 & $-$1.4~/~$-$0.7    \\

$b.$ $\rightarrow$ w/o Conv.                  & 77.6~/~92.9 & $-$0.5~/~$-$0.7   \\

$c.$ $\rightarrow$ ReLU                       & 77.7~/~93.4 & $-$0.4~/~$-$0.2   \\

$d.$ $\rightarrow$ single $\gamma,\beta$      & 77.6~/~93.3 & $-$0.5~/~$-$0.3     \\
\hline
\end{tabular}
\label{tab:configuration}
\end{center}
\vspace{-0.3cm}
\end{table}

\textbf{Hyper-parameter $r$}.
We also evaluate the different group division strategy in the first ``Conv.'' of Fig.~\ref{sec:csnlayer} through controlling the hyper-parameter $r$.
Although the total numbers of parameters in ``Conv.'' layer are the same by using distinct $r$, the reduced feature dimensions are different, leading to the different computational complexity, \ie the larger $r$, the smaller computation cost in the subsequent block.
Table~\ref{tab:hyperpara2} shows the top-1 accuracy on ImageNet by using EN-ResNet50 with different group division in the first ``Conv.'' shown in Fig.~\ref{fig:CSN}.
All of the configurations achieve higher performance than SN.
With the value of $r$ growths, the performance of EN-ResNet50 increases stably expect $64$, which equals to the smallest number of channels in ResNet50.
These results indicate that feature dimension reduction benefits to the performance increment. However, such advantage may disappear if the reduction rate equals to the smallest number of channels.

\textbf{Other Configurations}.
We replace the other components in the EN layer to verify their effectiveness.
The configurations for comparison are as follows. ($a$) A 2-layer multi-layer perceptron (MLP) is used to replace the designed important ratio calculation module in Fig.~\ref{fig:CSN}.
The MLP reduces the feature dimension to $1/32$ in the first layer followed by an activation function, and then reduce the dimension to the number of important ratios in the second layer.
($b$) The ``Conv.'' operation in the Fig.~\ref{fig:CSN} are omitted and pairwise correlations $\bm{v}_n$ in Sec.~\ref{sec:csnlayer} `step(2)' are directly computed.
($c$) The Tanh activation function in the top blue block of Fig.~\ref{fig:CSN} is replaced with ReLU.
($d$) Instead of multiple $\gamma,\beta$ in Eqn.~\eqref{eq:csn} (\ie each $\gamma,\beta$ is corresponding to one normalizer), single $\gamma,\beta$ are adopted.
Table~\ref{tab:configuration} reports the comparisons of proposed EN with different internal configuration.
According to the results, the current configuration of EN achieves the best performance compared with the other variants.
It is worthy to note that we find the output of 2-layer MLP changing dramatically in the training phase (\ie important ratios), making the distribution of feature maps at different iterations changed too much and leading to much poor accuracy.

%

\begin{figure}[t]
\centering
\subfigure[The average ratios on ImageNet validation set ]{
\begin{minipage}[t]{1\linewidth}
\centering
\includegraphics[width=1\linewidth]{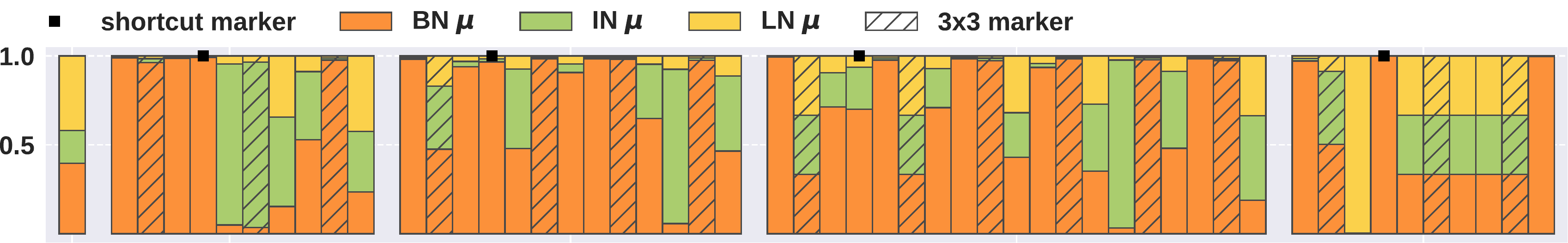}
\end{minipage}%
} \\
\vspace{-0.2cm}
\subfigure[The average ratios on Webvision validation set]{
\begin{minipage}[t]{1\linewidth}
\centering
\includegraphics[width=1\linewidth]{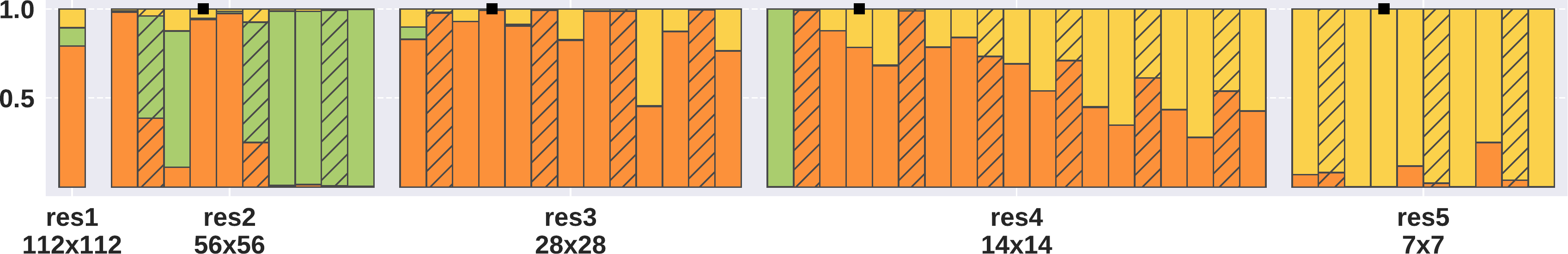}
\end{minipage}%
}%
\centering
\vspace{-0.3cm}
\caption{The averaged sample ratios in different layers of ResNet50 on ImageNet and Webvision validation set.
The $y$-axis denotes the important ratios of different normalizers after the softmax operation (\ie sum to $1$).
The $x$-axis shows different residual blocks of ResNet50 and the image resolution in each block is represented as well.
%
%
Different datasets learn distinct averaged ratios for different normalizers in different layers of the network.
}
\label{fig:DatasetRatioComparison}
\end{figure}

\subsection{Analysis of EN}



\textbf{Learning Dynamic of Ratios on Dataset.}
Since the parameters which are adopted to learn the important ratios $\bm{\lambda}$ in EN layer are initialized as $0$, the important ratios of each sample in each layer have uniform values ( \ie $1/3$ ) at the beginning of the model training.
In the training phase, the values of $\bm{\lambda}$ changes between $0$ and $1$.
We first investigate the averaged sample ratios in different layers of ResNet50 on ImageNet and Webvision validation set.
We use the optimized model to calculate the ratios of each sample in each layer, then the average ratios of each layer are calculated over all of the validation set.
According to Fig.~\ref{fig:DatasetRatioComparison}, once the training dataset is determined,
the learned averaged ratios are usually distinct for different datasets.

\begin{figure}
\begin{center}
\includegraphics[width=1.0\linewidth]{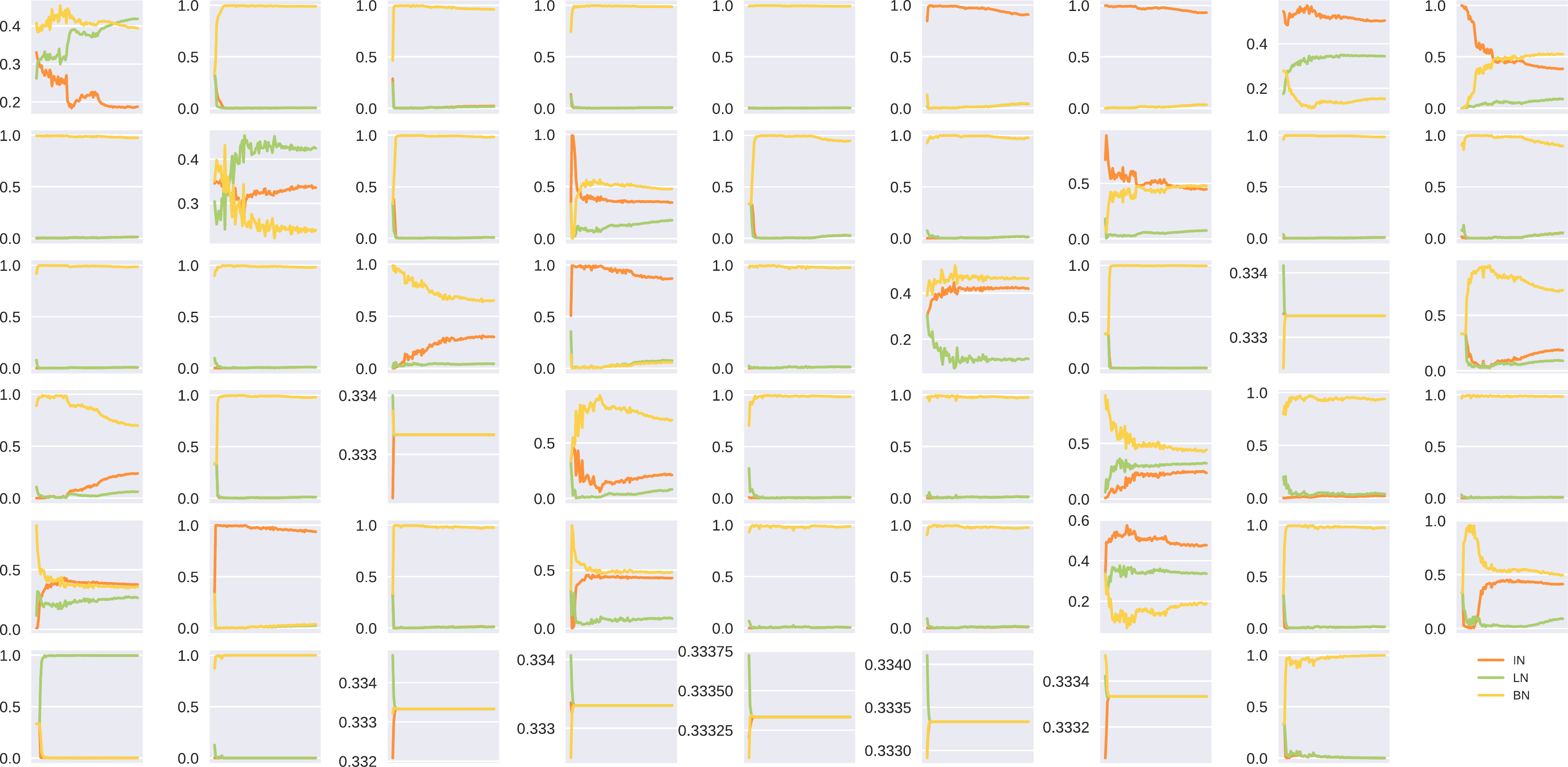}
\end{center}
\caption{The visualization of averaged sample ratios in $53$ normalization layers of EN-ResNet50 trained on ImageNet for 100 epoches.
The y-axis of each sub-figure denotes the important ratios of different normalizers.
The x-axis shows the different training epoches.
Zoom in three times for the best view.}
\label{fig:visualdataset}
\vspace{-0.6cm}
\end{figure}

To analysis the changes of ratios in the training process,
Fig.~\ref{fig:visualdataset} plots the leaning dynamic of ratios of $100$ epochs for $53$ normalization layers in ResNet50 .
Each value of ratios are averaged over all of the samples in ImageNet validation set.
From the perspective of the entire dataset, the changes of ratios in each layer of EN are similar to those in SN, whose values have smooth fluctuation in the training phase,
implying that distinct layers may need their own preference of normalizers to optimize the model in different epochs.

\begin{figure}
\begin{center}
\includegraphics[width=1.0\linewidth]{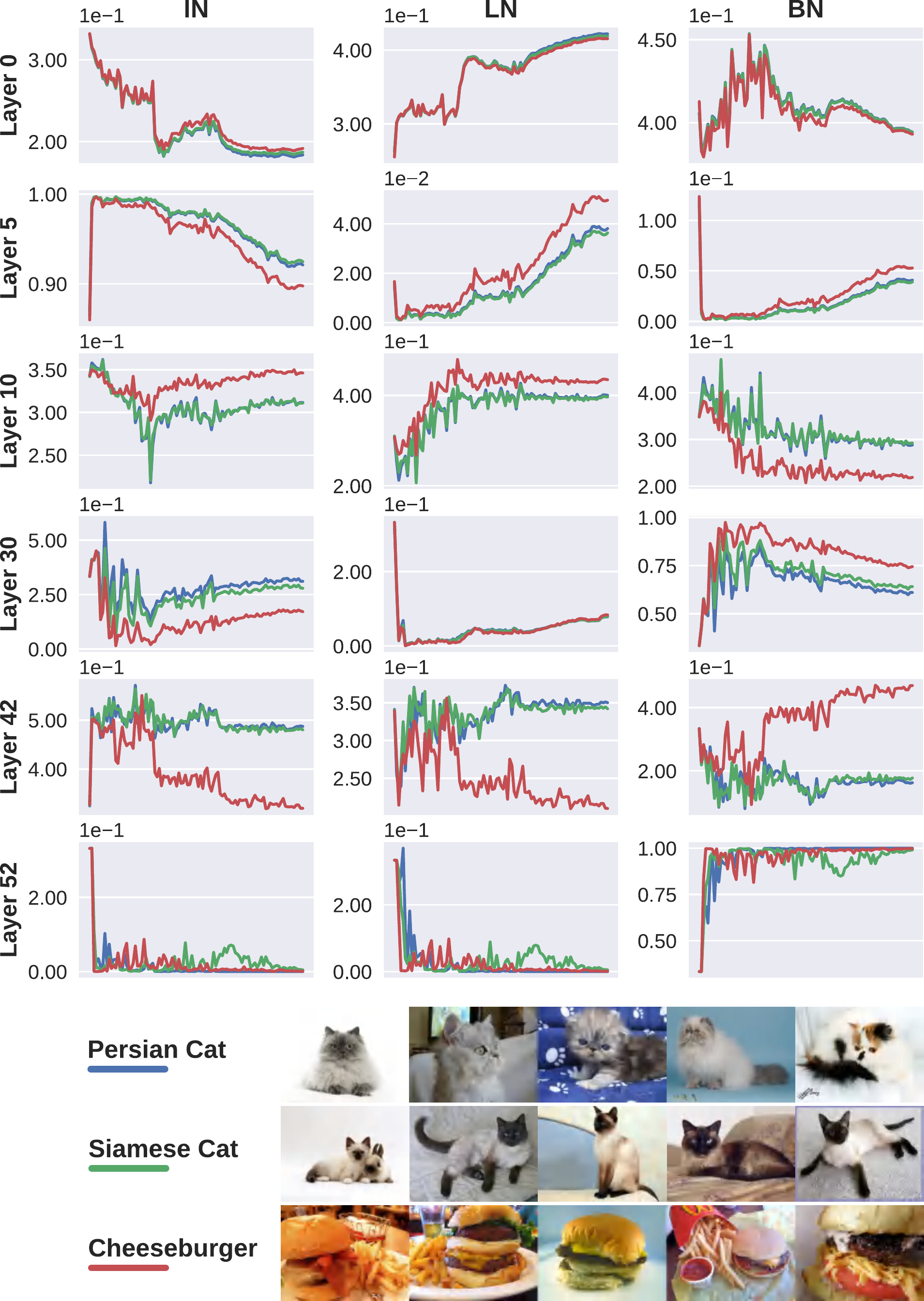}
\end{center}
\caption{The visualization of the important ratios of 3 categories (\ie Persian cat, Siamese cat and Cheeseburger ) in 6 different layers of ResNet50. Each column indicates one of the normalizers.  }
\label{fig:visualclass}
\vspace{-0.4cm}
\end{figure}

\begin{figure}
\begin{center}
\includegraphics[width=1.0\linewidth]{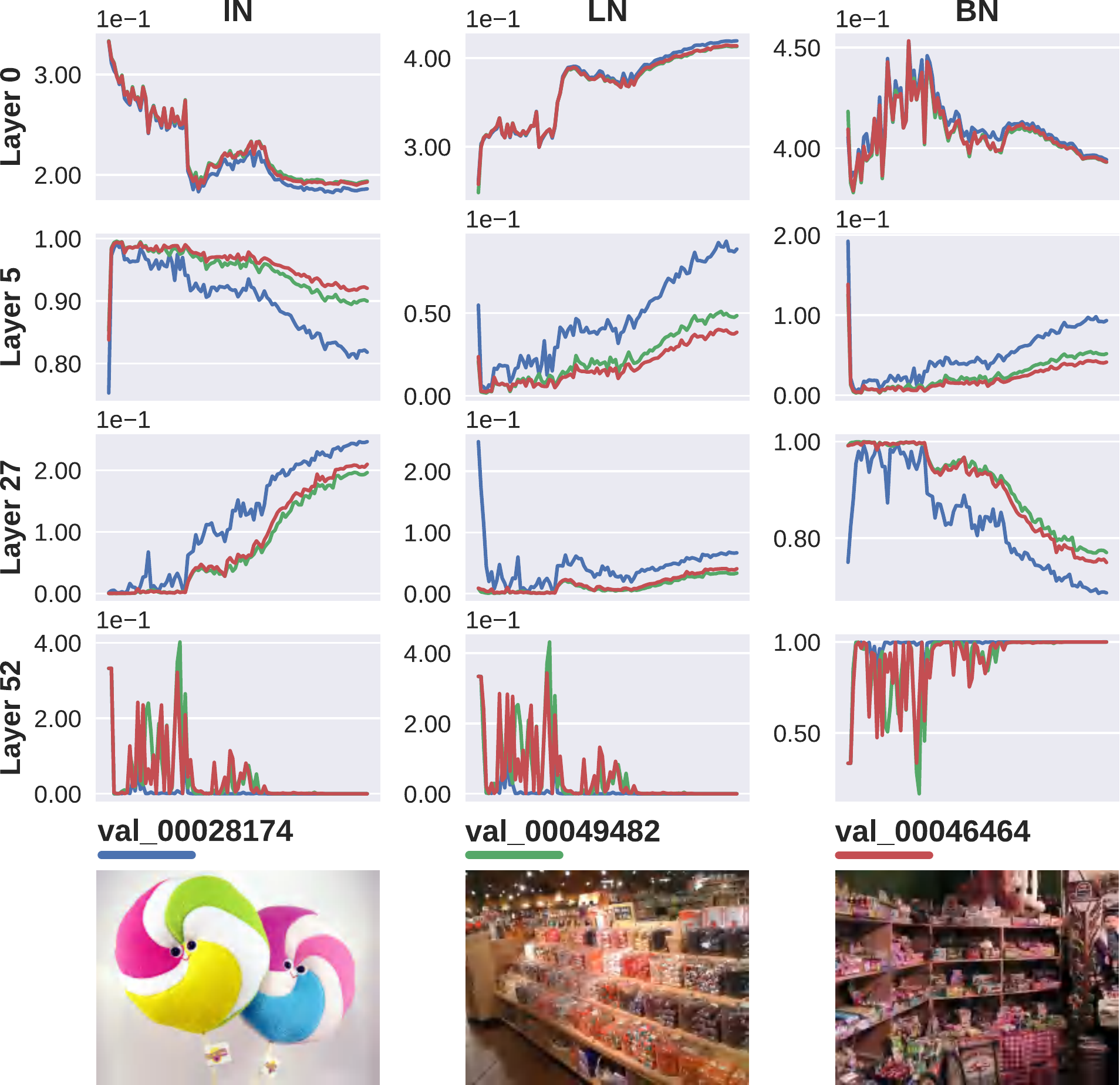}
\end{center}
\caption{The visualization of the important ratios of 3 samples selected from Confectionery class in different layers of ResNet50.  }
\label{fig:visualsample}
\vspace{-0.4cm}
\end{figure}

\textbf{Learning Dynamic of Ratios on Classes and Images.}
One advantage of EN compared with SN is able to learn important ratios to adaptive to different exemplars.
To illustrate such benefit of EN, we further plot the averaged important ratios of different classes (\ie w/ and w/o similar appearance)  in different layers in Fig.~\ref{fig:visualclass},
as well as the important ratios of various image samples in different layers in Fig.~\ref{fig:visualsample}.
We have the following observations.

(1) Different classes learn their own important ratios in different layers.
However, once the neural network is optimized on a certain dataset (\eg ImageNet), the trend of the ratio changes of are similar in different epochs.
For example, in Fig.~\ref{fig:visualclass}, since the Persian cat and Siamese cat have a similar appearance, their leaned ratio curves are very close and even coincident in some layers, \eg Layer5 and Layer 10.
While the ratio curves from the class of Cheeseburger are far away from the above two categories.
But in most layers, the ratio changes of different normalizers are basically the same, only have the numerical nuances.


(2) For the images with the same class index but various appearances, their learned ratios could also be distinct in different layers.
Such cases are shown in Fig.~\ref{fig:visualsample}.
All of the images are sampled from confectionery class but with various appearance, \eg the exemplar of confectionery and shelves for selling candy.
According to Fig.~\ref{fig:visualsample}, different images from the same category also obtained different ratios in bottom, middle and top normalization layers.


\begin{figure*}[t]

\begin{center}
\centering
\includegraphics[width=\linewidth]{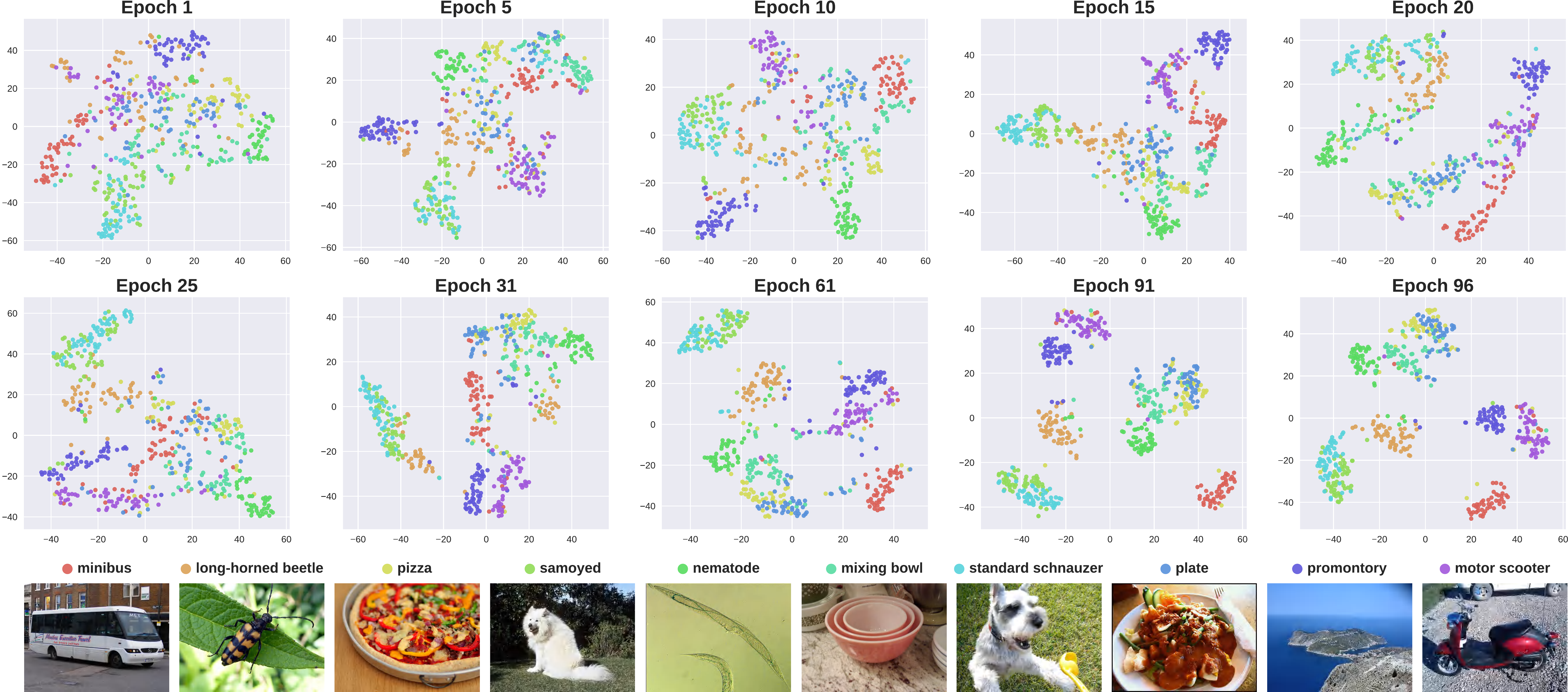}
\end{center}

\caption{The visualization of sample clustering on $10$ categories of ImageNet validation set by using learned important ratios.
For each epoch, the concatenated important ratios of each sample are reduced to $2$-D by using t-SNE~\cite{J:t-sne}.
The corresponding epoch number is shown in the top of each sub-figure.
Note that the $31$, $61$ and $91$ epoch are three epochs after the learning rate decay.
Each color is corresponding to a specific semantic category.
Better view in color with zooming in.}
\label{fig:clustering}
\end{figure*}

\section{Conclusion}

In this paper, we propose Exemplar Normalization to learn the linear combination of different normalizers with a sample-based manner in a single layer.
We show the effectiveness of EN on various computer vision tasks,
such as classification, detection and segmentation,
demonstrate its superior learning and generalization ability than static learning-to-normalize method such as SN.
In addition, the interpretable visualization of learned important ratios reveals the properties of classes and datasets.
The future work will explore EN in more intelligent tasks.
%
%
In addition, the task-oriented constraint on the important ratios will also be a potential research direction.

\noindent
\textbf{Acknowledgement} This work was partially supported by No. 2018YFB1800800, Open Research Fund from Shenzhen Research Institute of Big Data No. 2019ORF01005, 2018B030338001, 2017ZT07X152, ZDSYS201707251409055, HKU Seed Fund for Basic Research and Start-up Fund.

 \section*{Appendix}

\textbf{Sample Clustering via Important Ratios.}
Exemplar Normalization (EN) provides another perspective to understand the structure information in CNNs.
To further analyze the effect of proposed EN on capturing the semantic information,
we concatenate the learned important ratios in all of the EN layers for the input images and adopt t-Distributed Stochastic Neighbor Embedding (t-SNE)~\cite{J:t-sne} to reduce the dimensions to 2-D.
The visualization of these samples are shown in Fig.~\ref{fig:clustering}.

In practice, we train EN-ResNet50 on the ImageNet~\cite{C:ImageNet} training set.
The normalizer pool used in EN is $\{$ IN, LN, BN $\}$.
Then we randomly select $10$ categories from ImageNet validation set to visualize the sample distribution.
For each categories, all of the validation samples are used (\ie $50$ samples per category).
The name of the selected categories and related exemplary images are present at bottom of Fig.~\ref{fig:clustering}.
To visualize each sample, we extract and concatenate its important ratios from all of the EN layer in EN-ResNet50.
Thus the dimension of concatenated important ratios is $53 \times 3 = 159$.
Then we use the open source of t-SNE\footnote{\url{https://lvdmaaten.github.io/tsne/}} to reduce the   dimension from $159$ to $2$ to visualize the sample distribution.
We select $10$ typical training epochs to show the clustering dynamic in the training phase.

According to Fig.~\ref{fig:clustering}, we have the following observations.
(1) The learned important ratios can be treated as one type of structure information to realize \textbf{semantic preservation}.
When the model converges, \ie at $96$ epoch, the samples with the same label are grouped into the same cluster.
It further demonstrates different categories tend to select different normalizers to further improve their representation abilities, as well as the prediction accuracy of the model.
(2) The learned important ratios in EN also makes \textbf{appearance embedding} possible.
For example, the \texttt{samoyed} and \texttt{standard schnauzer} have the same father category according to the WordNet\footnote{\url{https://wordnet.princeton.edu/}} hierarchy and the samples in these two categories share the same appearance.
Thus, the distance between the corresponding two clusters are small.
The same result also achieves in category \texttt{pizza} and \texttt{plate}.
But cluster \texttt{samoyed} is far away from cluster \texttt{pizza} since they provide great difference in appearance.
(3) We also investigate the \textbf{clustering dynamic} in Fig.~\ref{fig:clustering}.
We show the sample distributions in $10$ different epochs of training process.
In the beginning of the model training, all of the samples are uniform  distributed and none of semantic clusters are generated.
From $5$ epoch to $25$ epoch, the semantic clusters are generated rapidly along with the model optimization.
The semantic clusters are basically formed after $31$ epoch, which is the first epoch after the first time to decay the learning rate.
After that, the sample distribution are slightly adjusted in the rest epochs.

{\small
\bibliographystyle{ieee_fullname}
\bibliography{egbib}

\begin{thebibliography}{10}\itemsep=-1pt

\bibitem{A:LN}
Jimmy~Lei Ba, Jamie~Ryan Kiros, and Geoffrey~E. Hinton.
\newblock Layer normalization.
\newblock {\em arXiv:1607.06450}, 2016.

\bibitem{C:AANet}
Irwan Bello, Barret Zoph, Ashish Vaswani, Jonathon Shlens, and Quoc~V Le.
\newblock Attention augmented convolutional networks.
\newblock In {\em ICCV}, 2019.

\bibitem{J:deeplab}
Liang-Chieh Chen, George Papandreou, Iasonas Kokkinos, Kevin Murphy, and Alan~L
  Yuille.
\newblock Deeplab: Semantic image segmentation with deep convolutional nets,
  atrous convolution, and fully connected crfs.
\newblock {\em IEEE transactions on pattern analysis and machine intelligence},
  40(4):834--848, 2017.

\bibitem{C:Anet}
Yunpeng Chen, Yannis Kalantidis, Jianshu Li, Shuicheng Yan, and Jiashi Feng.
\newblock A\^{} 2-nets: Double attention networks.
\newblock In {\em NeurIPS}, 2018.

\bibitem{C:cityscape}
Marius Cordts, Mohamed Omran, Sebastian Ramos, Timo Rehfeld, Markus Enzweiler,
  Rodrigo Benenson, Uwe Franke, Stefan Roth, and Bernt Schiele.
\newblock The cityscapes dataset for semantic urban scene understanding.
\newblock In {\em CVPR}, 2016.

\bibitem{C:ImageNet}
Jia Deng, Wei Dong, Richard Socher, Li-Jia Li, Kai Li, and Li Fei-Fei.
\newblock Imagenet: A large-scale hierarchical image database.
\newblock In {\em CVPR}, 2009.

\bibitem{C:HighOrder}
Zilin Gao, Jiangtao Xie, Qilong Wang, and Peihua Li.
\newblock Global second-order pooling convolutional networks.
\newblock In {\em CVPR}, 2019.

\bibitem{C:CurriculumNet}
Sheng Guo, Weilin Huang, Haozhi Zhang, Chenfan Zhuang, Dengke Dong, Matthew~R
  Scott, and Dinglong Huang.
\newblock Curriculumnet: Weakly supervised learning from large-scale web
  images.
\newblock In {\em ECCV}, 2018.

\bibitem{C:Resnet}
Kaiming He, Xiangyu Zhang, Shaoqing Ren, and Jian Sun.
\newblock Deep residual learning for image recognition.
\newblock In {\em CVPR}, 2016.

\bibitem{C:ResV2}
Kaiming He, Xiangyu Zhang, Shaoqing Ren, and Jian Sun.
\newblock Identity mappings in deep residual networks.
\newblock In {\em ECCV}, 2016.

\bibitem{C:SENet}
Jie Hu, Li Shen, and Gang Sun.
\newblock Squeeze-and-excitation networks.
\newblock In {\em CVPR}, 2018.

\bibitem{C:CWN}
Lei Huang, Xianglong Liu, Yang Liu, Bo Lang, and Dacheng Tao.
\newblock Centered weight normalization in accelerating training of deep neural
  networks.
\newblock In {\em ICCV}, 2017.

\bibitem{C:BRN}
Sergey Ioffe.
\newblock Batch renormalization: Towards reducing minibatch dependence in
  batch-normalized models.
\newblock In {\em NeurIPS}, 2017.

\bibitem{C:BN}
Sergey Ioffe and Christian Szegedy.
\newblock Batch normalization: Accelerating deep network training by reducing
  internal covariate shift.
\newblock In {\em ICML}, 2015.

\bibitem{C:DynamicFilter}
Xu Jia, Bert De~Brabandere, Tinne Tuytelaars, and Luc~V Gool.
\newblock Dynamic filter networks.
\newblock In {\em NeurIPS}, 2016.

\bibitem{J:generalizingPooling}
Chen-Yu Lee, Patrick~W Gallagher, and Zhuowen Tu.
\newblock Generalizing pooling functions in convolutional neural networks:
  Mixed, gated, and tree.
\newblock In {\em Artificial intelligence and statistics}, 2016.

\bibitem{A:PN}
Boyi Li, Felix Wu, Kilian~Q. Weinberger, and Serge~J. Belongie.
\newblock Positional normalization.
\newblock {\em arXiv:1907.04312}, 2019.

\bibitem{A:webvision}
Wen Li, Limin Wang, Wei Li, Eirikur Agustsson, and Luc Van~Gool.
\newblock Webvision database: Visual learning and understanding from web data.
\newblock {\em arXiv:1708.02862}, 2017.

\bibitem{A:AN}
Xilai Li, Wei Sun, and Tianfu Wu.
\newblock Attentive normalization.
\newblock {\em arXiv:/1908.01259}, 2019.

\bibitem{A:StreamNrom}
Qianli Liao, Kenji Kawaguchi, and Tomaso Poggio.
\newblock Streaming normalization: Towards simpler and more
  biologically-plausible normalizations for online and recurrent learning.
\newblock {\em arXiv:1610.06160}, 2016.

\bibitem{C:MN}
Iain~Murray Lucas~Deecke and Hakan Bilen.
\newblock Mode normalization.
\newblock In {\em ICLR}, 2019.

\bibitem{A:DoNorm}
Ping Luo, Zhanglin Peng, Jiamin Ren, and Ruimao Zhang.
\newblock Do normalization layers in a deep convnet really need to be distinct?
\newblock {\em arXiv:1811.07727}, 2018.

\bibitem{C:SN}
Ping Luo, Jiamin Ren, Zhanglin Peng, Ruimao Zhang, and Jingyu Li.
\newblock Differentiable learning-to-normalize via switchable normalization.
\newblock In {\em ICLR}, 2019.

\bibitem{J:SN}
Ping Luo, Ruimao Zhang, Jiamin Ren, Zhanglin Peng, and Jingyu Li.
\newblock Switchable normalization for learning-to-normalize deep
  representation.
\newblock {\em {IEEE} Trans. Pattern Anal. Mach. Intell.}, 2019.

\bibitem{C:DN}
Ping Luo, Peng Zhanglin, Shao Wenqi, Zhang Ruimao, Ren Jiamin, and Wu Lingyun.
\newblock Differentiable dynamic normalization for learning deep
  representation.
\newblock In {\em ICML}, 2019.

\bibitem{C:ShuffleNetV2}
Ningning Ma, Xiangyu Zhang, Hai{-}Tao Zheng, and Jian Sun.
\newblock Shufflenet {V2:} practical guidelines for efficient {CNN}
  architecture design.
\newblock In {\em ECCV}, 2018.

\bibitem{J:t-sne}
Laurens van~der Maaten and Geoffrey Hinton.
\newblock Visualizing data using t-sne.
\newblock {\em Journal of machine learning research}, 9(Nov):2579--2605, 2008.

\bibitem{C:SpectralNorm}
Takeru Miyato, Toshiki Kataoka, Masanori Koyama, and Yuichi Yoshida.
\newblock Spectral normalization for generative adversarial networks.
\newblock In {\em ICLR}, 2018.

\bibitem{C:BIN}
Hyeonseob Nam and Hyo{-}Eun Kim.
\newblock Batch-instance normalization for adaptively style-invariant neural
  networks.
\newblock In {\em NeurIPS}, 2018.

\bibitem{C:IBN}
Xingang Pan, Ping Luo, Jianping Shi, and Xiaoou Tang.
\newblock Two at once: enhancing learning and generalization capacities via
  ibn-net.
\newblock In {\em ECCV}, 2018.

\bibitem{C:SW}
Xingang Pan, Xiaohang Zhan, Jianping Shi, Xiaoou Tang, and Ping Luo.
\newblock Switchable whitening for deep representation learning.
\newblock In {\em ICCV}, 2019.

\bibitem{A:CBN}
Ethan Perez, Harm de Vries, Florian Strub, Vincent Dumoulin, and Aaron~C.
  Courville.
\newblock Learning visual reasoning without strong priors.
\newblock {\em arXiv:1707.03017}, 2017.

\bibitem{C:WN}
Tim Salimans and Durk~P Kingma.
\newblock Weight normalization: A simple reparameterization to accelerate
  training of deep neural networks.
\newblock In {\em NIPS}, 2016.

\bibitem{C:SSN}
Wenqi Shao, Tianjian Meng, Jingyu Li, Ruimao Zhang, Yudian Li, Xiaogang Wang,
  and Ping Luo.
\newblock Ssn: Learning sparse switchable normalization via sparsestmax.
\newblock In {\em CVPR}, 2019.

\bibitem{A:CENet}
Wenqi Shao, Shitao Tang, Xingang Pan, Ping Tan, Xiaogang Wang, and Ping Luo.
\newblock Channel equilibrium networks for learning deep representation.
\newblock {\em arXiv:2003.00214}, 2020.

\bibitem{C:Inception-v3}
Christian Szegedy, Vincent Vanhoucke, Sergey Ioffe, Jon Shlens, and Zbigniew
  Wojna.
\newblock Rethinking the inception architecture for computer vision.
\newblock In {\em CVPR}, 2016.

\bibitem{A:IN}
Dmitry Ulyanov, Andrea Vedaldi, and Victor Lempitsky.
\newblock Instance normalization: The missing ingredient for fast stylization.
\newblock {\em arXiv:1607.08022}, 2016.

\bibitem{C:BKN}
Guangrun Wang, Jiefeng Peng, Ping Luo, Xinjiang Wang, and Liang Lin.
\newblock Batch kalman normalization: Towards training deep neural networks
  with micro-batches.
\newblock 2018.

\bibitem{C:ECANet}
Qilong Wang, Banggu Wu, Pengfei Zhu, Peihua Li, Wangmeng Zuo, and Qinghua Hu.
\newblock Eca-net: Efficient channel attention for deep convolutional neural
  networks.
\newblock In {\em CVPR}, 2020.

\bibitem{C:GN}
Yuxin Wu and Kaiming He.
\newblock Group normalization.
\newblock In {\em ECCV}, 2018.

\bibitem{J:PDN}
Ruimao Zhang, Wei Yang, Zhanglin Peng, Pengxu Wei, Xiaogang Wang, and Liang
  Lin.
\newblock Progressively diffused networks for semantic visual parsing.
\newblock {\em Pattern Recognition}, 90:78--86, 2019.

\bibitem{C:L2G}
Zhaoyang Zhang, Jingyu Li, Wenqi Shao, Zhanglin Peng, Ruimao Zhang, Xiaogang
  Wang, and Ping Luo.
\newblock Differentiable learning-to-group channels via groupable convolutional
  neural networks.
\newblock In {\em ICCV}, 2019.

\bibitem{C:ADE20K}
Bolei Zhou, Hang Zhao, Xavier Puig, Sanja Fidler, Adela Barriuso, and Antonio
  Torralba.
\newblock Scene parsing through {ADE20K} dataset.
\newblock In {\em CVPR}, 2017.

\end{thebibliography}
}

\end{document}